\documentclass[lettersize,journal]{IEEEtran}
\usepackage{amsmath,amsfonts}
\usepackage{algorithmic}
\usepackage{algorithm}
\usepackage{array}
\usepackage[caption=false,font=normalsize,labelfont=sf,textfont=sf]{subfig}
\usepackage{textcomp}
\usepackage{stfloats}
\usepackage{url}
\usepackage{verbatim}
\usepackage{graphicx}
\usepackage{cite}
\usepackage{color}
\definecolor{lightblue}{RGB}{0, 110, 204}
\usepackage{amsfonts}
\usepackage[table]{xcolor}
\usepackage{graphicx}
\usepackage{amsmath}
\usepackage{amsthm}
\usepackage[
pagebackref=true,
breaklinks=true,
letterpaper=true,
citecolor=lightblue,
colorlinks,
bookmarks=false
]{hyperref}
\usepackage{bbding}
\usepackage{wrapfig}
\usepackage{subfig}
\usepackage{multirow}
\usepackage{booktabs}       

\begin{document}


\def\eg{\textit{e.g.}}
\def\ie{\textit{i.e.}}
\def\etal{\textit{et al.}}
\def\methodname{XXX}

\title{CustomVideo: Customizing Text-to-Video Generation with Multiple Subjects}


\author{Zhao Wang, \IEEEmembership{Member, IEEE}, Aoxue Li, Lingting Zhu, Yong Guo, \\Qi Dou, \IEEEmembership{Senior Member, IEEE}, Zhenguo Li, \IEEEmembership{Member, IEEE}
\thanks{Zhao Wang, and Qi Dou are with Department of Computer Science and Engineering, The Chinese University of Hong Kong. (Email: \{zwang21@cse., qidou@\}cuhk.edu.hk)}
\thanks{Aoxue Li, Yong Guo, and Zhenguo Li are with Huawei Noah Ark's Lab. (Email: lax@pku.edu.cn, guoyongcs@gmail.com, li.zhenguo@huawei.com)}
\thanks{Lingting Zhu is with Department of Statistics and Actuarial Science, The University of Hong Kong. (Email: ltzhu99@connect.hku.hk)}
}

\markboth{Journal of \LaTeX\ Class Files,~Vol.~14, No.~8, August~2021}%
{Shell \MakeLowercase{\textit{et al.}}: A Sample Article Using IEEEtran.cls for IEEE Journals}





\maketitle

\begin{figure*}[t]
  \centering
  \vspace{-6mm}
   \includegraphics[width=\linewidth]{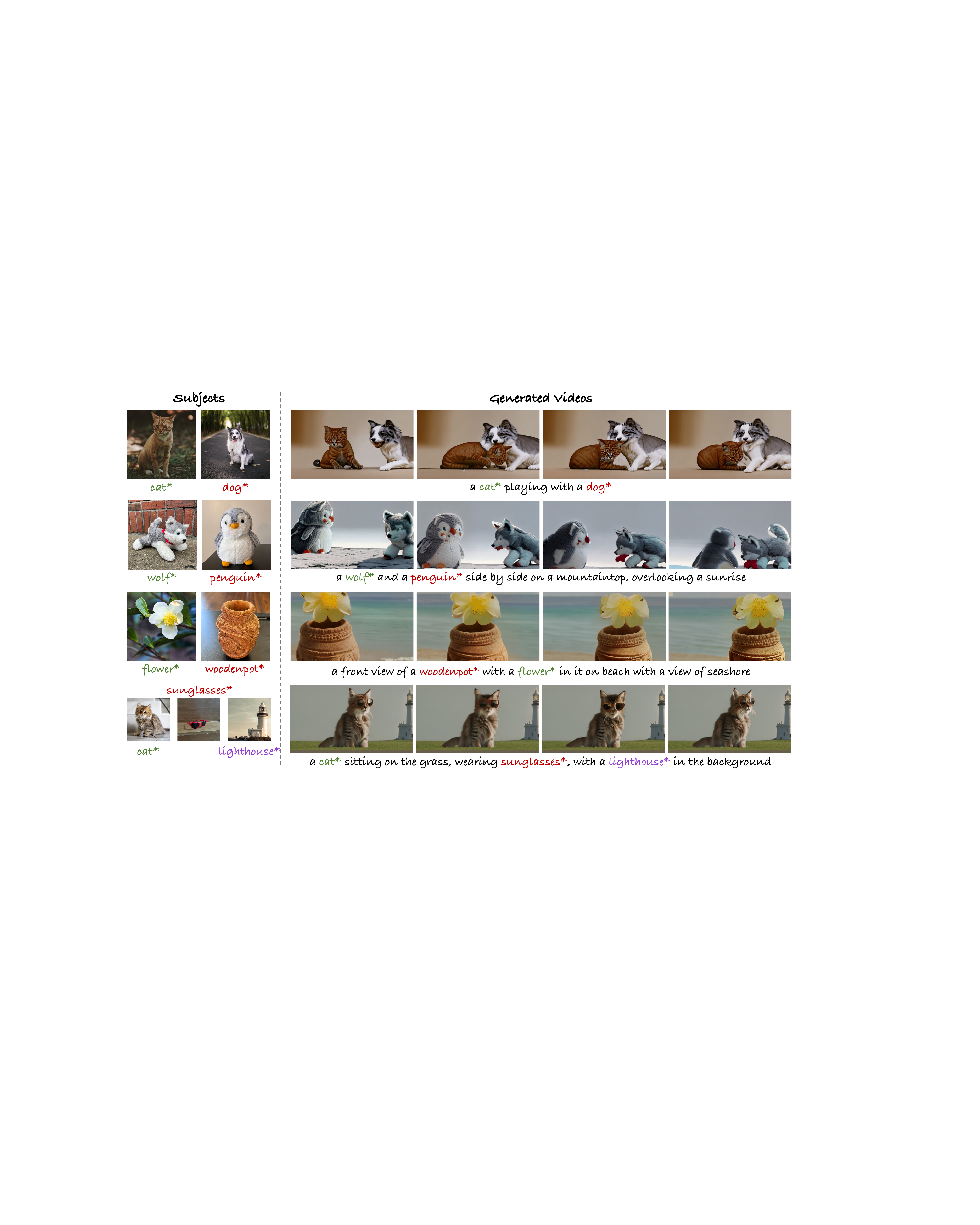}
   \vspace{-6mm}
   \caption{
   \textbf{Customized text-to-video generation results} of our proposed CustomVideo given multiple subjects (left) and text prompts (below). Our approach can disentangle highly similar subjects, \eg, \emph{cat} v.s. \emph{dog}, simultaneously preserving the fidelity of subjects and smooth motions.
   }
   \label{fig:teaser}
\end{figure*}

\begin{abstract}

Customized text-to-video generation aims to generate high-quality videos guided by text prompts and subject references. Current approaches for personalizing text-to-video generation suffer from tackling multiple subjects, which is a more challenging and practical scenario. In this work, our aim is to promote multi-subject guided text-to-video customization. We propose CustomVideo, a novel framework that can generate identity-preserving videos with the guidance of multiple subjects. To be specific, firstly, we encourage the co-occurrence of multiple subjects via composing them in a single image. Further, upon a basic text-to-video diffusion model, we design a simple yet effective attention control strategy to disentangle different subjects in the latent space of diffusion model. Moreover, to help the model focus on the specific area of the object, we segment the object from given reference images and provide a corresponding object mask for attention learning. Also, we collect a multi-subject text-to-video generation dataset as a comprehensive benchmark. Extensive qualitative, quantitative, and user study results demonstrate the superiority of our method compared to previous state-of-the-art approaches. The project page is \url{https://kyfafyd.wang/projects/customvideo}.
\end{abstract}


\section{Introduction}

Text-to-video (T2V) generation \cite{wang2023modelscope,zeng2023make,blattmann2023stable,kondratyuk2023videopoet,chen2024videocrafter2} has achieved fantastic progress taking advantages of diffusion models \cite{ho2020denoising,song2020denoising,lu2022dpm}.
Recently, artists have dreamed of generating videos with their own belongings, \eg, pets, which directs a new research direction named customized T2V generation.
Although existing methods \cite{zhao2023videoassembler,jiang2024videobooth,wei2024dreamvideo} have been proposed to generate videos from a single object, tackling multiple objects still remains a difficult scenario.
The key challenge is to ensure the co-occurrence of multiple objects in the generated video and retain their corresponding identities.

A recent work, VideoDreamer \cite{chen2023videodreamer}, proposes disen-mix finetuning and human-in-the-loop re-finetuning strategies based on Stable Diffusion \cite{rombach2022high}, aiming to generate videos from multiple subjects. 
However, VideoDreamer falls short in guaranteeing the co-occurrence of multiple subjects and disentangling similar subjects due to its inconsistent object mixing strategy. 
As shown in Fig.\ref{fig:sota_comparison}, both the car and cat can not be consistently generated across all frames.
Another straight forward solution for multi-subject driven T2V generation is by combining multi-subject driven Text-to-Image (T2I) and Image-to-Video (I2V) models.
However, this naive approach fails from two aspects, including inaccurate T2I customization and lack of motion (see Fig.\ref{fig:sota_comparison} `DisenDiff \cite{zhang2024attention} + SVD \cite{blattmann2023stable}').

In contrast, in our approach, we ensure the co-occurrence of multiple objects during model training for the first time in multi-subject driven T2V customization, which encourages the model to capture the presence of different subjects simultaneously, thereby facilitating co-occurrence during inference. 
Additionally, we propose an attention control mechanism to disentangle multiple subjects during training, effectively guiding the model to focus on the corresponding subject area while disregarding irrelevant parts of the image. 
To facilitate this process, we incorporate a ground truth object mask, obtained through segmentation either from a model like SAM \cite{kirillov2023segany} or provided by human annotators, as supervision during optimization. 
Our attention mechanism consists of two key designs that contribute to the disentanglement of subjects. 
Firstly, we highlight the corresponding subject area using the ground truth object mask on the cross-attention map, promoting better preservation of subject identity. 
Secondly, we optimize the cross-attention map towards a slight negative value, excluding the desired subject and mitigating the influence of irrelevant area in the input image. 
To comprehensively evaluate our proposed approach, we have curated a diverse dataset covering a wide range of categories.
Through extensive experiments on this benchmark dataset, we provide qualitative, quantitative, and user study results that demonstrate the superiority of our method in generating high-quality videos with customized subjects.
In summary, our contributions are as follows:

\begin{itemize}
    \item We propose CustomVideo, a novel multi-subject driven T2V generation framework, powered by a simple yet effective co-occurrence and attention control mechanism.
    \item We collect a multi-subject T2V dataset and build it as a comprehensive benchmark, which covers a wide range of subject categories and diverse subject pairs over them.
    \item Our method consistently outperforms previous state-of-the-art (SOTA) approaches in qualitative, quantitative and human preference evaluation, \eg, 11.99\% \emph{CLIP Image Alignment} and 23.39\% \emph{DINO Image Alignment} better than the previous best method.
\end{itemize}




\section{Related Work}

\paragraph{Text-to-Video Generation}
Text-to-video generation has made significant advancements in recent years \cite{chen2023videocrafter1,duan2023diffsynth,wang2023modelscope}. 
Early approaches \cite{balaji2019conditional,skorokhodov2022stylegan,hong2022cogvideo,villegas2023phenaki,zhu2023motionvideogan} employed GANs \cite{goodfellow2014generative} and VQ-VAE \cite{van2017neural}, while more recent works have explored diffusion models to generate high-quality videos \cite{he2022latent,wang2023lavie,zhang2023show,zhang2023i2vgen,hu2024benchmark,zhao2024ta2v,koksal2023controllable}. 
Make-A-Video \cite{singer2023makeavideo} utilizes a pre-trained image diffusion model with additional temporal attention finetuning. 
VideoLDM \cite{blattmann2023align} introduces a multi-stage alignment approach in the latent space to generate high-resolution and temporally consistent videos. 
Other methods \cite{khachatryan2023text2video,guo2023animatediff} generate videos with an image as the first frame and randomly initialized subsequent frames. 
To enhance controllability, VideoComposer \cite{wang2023videocomposer} incorporates additional guidance signals, such as depth maps, to produce desired videos alongside text inputs.
VideoDirectorGPT \cite{lin2023videodirectorgpt} aims to control the video generation with the guidance from different scenes and specific layouts generated by GPT4 in the temporal axis.
GEST \cite{masala2023explaining} models the representation between the text and video via a graph of events in the spatial and temporal space, in which the video is generated following a timeline.
Tune-A-Video \cite{wu2023tune} proposes a temporal self-attention module that fine-tunes a pre-trained image diffusion model, achieving successful generation of videos with specific text guidance. 
Furthermore, diffusion-based video-to-video editing approaches \cite{qi2023fatezero,geyer2023tokenflow} have also been proposed for practical usages.

\paragraph{Subject-driven Customization}
There has been a growing interest in customizing pre-trained image and video diffusion models for personalized generation \cite{gal2023an,kumari2023multi,avrahami2023break,ma2023subject,jiang2024animediff,liu2023cones,liu2023cones2,wei2024dreamvideo}. 
Customization involves generating images and videos with specific subjects, typically based on a few reference images. 
For image diffusion customization, Textual Inversion \cite{gal2023an} represents a specific object as a learnable text token using only a few reference images. 
This learned text token can then be integrated into a sentence to generate personalized images during inference. 
Additionally, DreamBooth \cite{ruiz2023dreambooth} fine-tunes the weights of the diffusion model to improve fidelity in image generation. 
Several works \cite{xiao2023fastcomposer,zhang2024attention,wei2024mm,jiang2024mc} have explored personalized image diffusion with multiple subjects, focusing on parameter-efficient finetuning and text embedding learning.
While there have been initial attempts to customize video diffusion, such as VideoAssembler \cite{zhao2023videoassembler}, VideoBooth \cite{jiang2024videobooth}, and ID-Animator \cite{he2024id}, which use reference images to personalize the video diffusion model while preserving subject identity, and DreamVideo \cite{wei2024dreamvideo}, which decouples the learning process for subject and motion customization, these methods are designed for single objects and cannot handle multiple subjects when given. 
A recent work, VideoDreamer \cite{chen2023videodreamer}, proposes multi-subject driven video customization through disen-mix finetuning strategy with LoRA \cite{hu2022lora}. 
However, the generated videos do not guarantee the co-occurrence of multiple subjects, or disentanglement of different subjects.
An alternative way is to combine multi-subject T2I customization and I2V models, such as DisenDiff \cite{zhang2024attention} + SVD-XT \cite{blattmann2023stable}.
However, such approach suffers from two fundamental limitations: 1) lack of spatial co-occurrence training: T2I models are typically trained on single-image supervision, lacking explicit signals to enforce the spatial co-occurrence of multiple subjects. In contrast, our CustomVideo trains on concatenated multi-subject images, forcing the model to learn their joint spatial distributions from the outset; 2) motion semantic gap: I2V models learn motion patterns from generic video data but lack subject-specific motion priors. For instance, in Figure \ref{fig:sota_comparison}, ``DisenDiff+SVD-XT'' generates static or disjointed motions for the car and barn, failing to capture dynamic interactions like ``stopping".
In this work, we propose a simple yet effective co-occurrence and attention mechanism that disentangles multiple subjects using masks as guidance while preserving the co-occurrence of subjects in the generated videos.

\begin{figure*}[t]
  \centering
   \includegraphics[width=\linewidth]{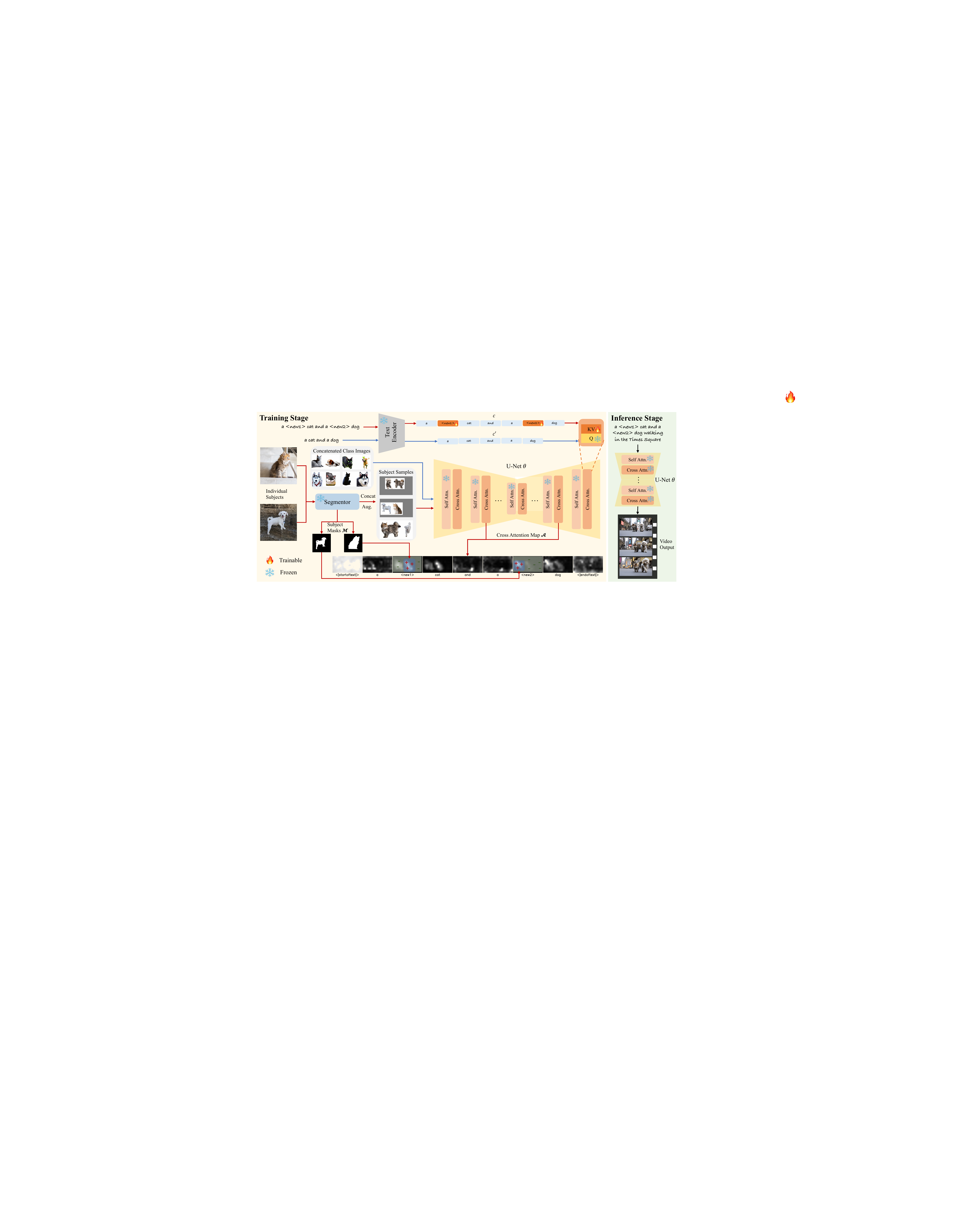}
   \vspace{-5mm}
   \caption{{\bf The overview of our proposed CustomVideo.} We propose a simple yet effective co-occurrence and attention control mechanism with mask guidance to preserve the fidelity of subjects for multi-subject driven text-to-video generation. 
   We also incorporate a class-specific prior preservation strategy to prevent language drift by feeding the model concatenated class images. During the training stage, only the key and value weights in the cross-attention layers are fine-tuned, together with a learnable word token for every subject. In the inference stage, given a text prompt integrated with learned word tokens, we can easily obtain high-quality videos with specific subjects. 
   }
   \label{fig:framework}
\end{figure*}

\section{Method}

\subsection{Preliminary: Text-to-Video}
\label{sec:t2v_preliminary}

Video diffusion models (VDMs) \cite{wang2023modelscope,duan2023diffsynth,chen2023videocrafter1} generate videos by gradually denoising a randomly sampled Gaussian noise $\epsilon$, following an iterative denoising process that resembles a reverse procedure of a fixed-length Markov Chain. 
This iterative denoising allows VDMs to capture the temporal dependencies presented in video data.
To be specific, a video diffusion model $\theta$ predicts the added noise at each timestep $t$ given a text condition $c$, where $t\in\{1, 2, \cdots, T\}$.
The training objective for this process can be expressed as a reconstruction loss:
\begin{equation}
\mathcal{L}_{recon}=\mathbb{E}_{\epsilon, \boldsymbol{z}, c, t}\left[\left\|\epsilon-\epsilon_\theta\left(\boldsymbol{z}_t, \mathcal{T}(c), t\right)\right\|_2^2\right],
\label{equ:t2v}
\end{equation}
where $\boldsymbol{z} \in \mathbb{R}^{B \times L \times H \times W \times D}$ is the latent code of the input video with batch size $B$, video length $L$, height $H$, width $W$, and latent dimension $D$.
$\epsilon_\theta$ is the noise prediction from the model.
$\mathcal{T}$ is a pre-trained text encoder.
$\boldsymbol{z}_t$ is obtained by adding noise to the ground truth $\boldsymbol{z}_0$ with $\boldsymbol{z}_t=\alpha_t \boldsymbol{z}_0+\sqrt{1-\alpha_t^2} \epsilon$, where $\alpha_t$ is a diffusion hyperparameter.
In this work, we utilize zeroscope \cite{zeroscope} T2V model as our base model, which is built upon a 3D U-Net, with spatial and temporal modeling for generating high-quality videos.

\subsection{CustomVideo with Multiple Subjects}
\label{sec:customvideo}

An overview of our proposed CustomVideo framework is illustrated in Fig. \ref{fig:framework}. 
For customization perspective, we introduce a new learnable word token for every subject, \eg, `$\textless$new1$\textgreater$' for cat and `$\textless$new2$\textgreater$' for dog, representing the corresponding identity.
During the training stage, we employ a concatenation technique to ensure the co-occurrence of multiple subjects. 
Specifically, we combine these subjects into a single image, facilitating model’s learning of multi-subject patterns.
To address the challenge of entanglement among highly similar subjects, we propose an attention control mechanism. 
This mechanism ensures that the learnable word tokens align with their corresponding regions on the cross-attention map (see Fig. \ref{fig:attention_visualization}). 
By achieving this alignment, we enable the model to disentangle different subjects and enhance the quality of the generated videos.
The training process focuses on training the subject-related learnable word tokens, as well as the key and value weights within the cross-attention layers of the U-Net architecture, adopting a parameter-efficient finetuning approach.
During the inference stage, users only need to provide a text prompt description integrated with the corresponding learned word tokens to generate high-quality videos aligned with their preferences.
In the following, we will delve into the detailed workings of CustomVideo.


\paragraph{Co-occurrence Control}
In the context of multi-subject driven T2V generation, ensuring that the model consistently learns to generate videos with multiple subjects is crucial. 
In a previous work, VideoDreamer \cite{chen2023videodreamer}, the authors proposed a disen-mix strategy for multi-subject driven generation by fine-tuning the model using both single subject images and concatenated images of multiple subjects.
However, we have identified that this mixing strategy can confuse the model due to the inconsistent number of subjects present in a single image. 
Consequently, it leads to unstable generation of videos with multiple subjects (see line `both single and concat' in Fig. \ref{fig:component_analysis}). In our approach, we have found that fine-tuning the model using only concatenated images of multiple subjects is sufficient to ensure the co-occurrence of multiple subjects in the generated videos.
Additionally, we have observed that providing clear subjects without background aids the model in learning the specific characteristics of the subjects more effectively. To achieve this, we perform background removal on the subject images. This can be accomplished manually or by employing an automatic tool such as the SAM model \cite{kirillov2023segany}.

\paragraph{Attention Control} 
Ensuring the co-occurrence of multiple subjects in generated videos is achieved through the above. 
However, a more challenging task is preserving the distinct characteristics of each subject without interference. 
Simply fine-tuning the model with concatenated images can lead to confusion among subject characteristics. 
As shown in Fig. \ref{fig:component_analysis} (line `w/o pos. attn.'), the generated cat predominantly resembles the shape of the provided dog sample, despite having similar color and texture to the provided cat. 
Thus, disentangling the multiple subjects becomes crucial to generate high-quality videos that faithfully represent each subject's characteristics.

In our approach, the learnable word tokens operate in the cross-attention layers of the diffusion model. 
To effectively regulate the subject learning process, we can directly leverage the cross-attention map.
As illustrated in Fig. \ref{fig:framework}, we employ an automatic segmentor to obtain the ground truth mask $\boldsymbol{\mathcal{M}}^p$ for each subject, which indicates the spatial position of the subjects. 
During the training stage, we extract the cross-attention map $\boldsymbol{\mathcal{A}}$ from the cross-attention layer using the activations of each word in the given text prompt.
For each learnable word token, such as `$\textless$new1$\textgreater$', we enhance the corresponding area on the cross-attention map $\boldsymbol{\mathcal{A}}$ by encouraging the alignment between the subject area and the ground truth subject mask with a loss function as the following:
\begin{equation}
    \mathcal{L}_{attn}^{pos}=\frac{1}{B}\sum_{i=1}^B\frac{1}{N} \sum_{j=1}^N\left\|\boldsymbol{\mathcal{A}}_{i,j}-\boldsymbol{\mathcal{M}}_{i,j}^p\right\|_2^2
    \label{equ:attn_pos}
\end{equation}
where $N$ is the number of subjects, $B$ is the training batch size, and the corresponding area of subject in the mask $\boldsymbol{\mathcal{M}}^p$ is filled by value 1 while the remaining area is filled by 0.
By employing this positive attention mechanism, the model is compelled to allocate more attention to the correct subject area, leading to an effective learning of the corresponding subject characteristics.


The positive-style attention mechanism mentioned above effectively enhances the learning of specific characteristics for each subject. However, there exist issues with the generated subjects, particularly when irrelevant area is present.
For example, in Fig. \ref{fig:component_analysis} (line `w/o neg. attn.'), we can observe that the generated dog's legs are affected by color information from the given cat, which is not desirable.
To address this problem, we introduce negative guidance by considering the area outside the subject. 
In addition to the positive guidance within the ground truth subject mask, we incorporate a small negative value, denoted as $\eta$, into the region outside the subject within the mask $\boldsymbol{\mathcal{M}}^p$. This modified mask, denoted as $\boldsymbol{\mathcal{M}}^{[p,n]}$, is then used to replace $\boldsymbol{\mathcal{M}}^p$ in the attention loss. 
So Eq. \eqref{equ:attn_pos} becomes
\begin{equation}
    \mathcal{L}_{attn}^{pos\_neg}=\frac{1}{B}\sum_{i=1}^B\frac{1}{N} \sum_{j=1}^N\left\|\boldsymbol{\mathcal{A}}_{i,j}-\boldsymbol{\mathcal{M}}_{i,j}^{[p,n]}\right\|_2^2.
    \label{equ:attn_pos_neg}
\end{equation}
By integrating negative guidance, we can alleviate the issue of irrelevant area and improve the fidelity of the generated subjects.


\subsection{Model Training and Inference}
\label{sec:training}

\paragraph{Training Strategy}

During training, we only fine-tune the weights of key and value in all of the cross-attention layers.
We use a more flexible approach by fine-tuning the model with subject images extended to single-frame videos as both input and supervision, rather than using subject videos.
This approach guarantees the co-occurrence of multiple subjects because subject images are much easier to obtain and smoother to concatenate compared to subject videos.
Moreover, fine-tuning with subject images saves lots of computation cost.
Inspired by previous T2I personalization \cite{ruiz2023dreambooth,kumari2023multi}, we conduct class-specific prior preservation to improve the diversity of generated videos and alleviate the issue of language drift. 
Specifically, we generate 200 class images with Stable Diffusion v2.1 \cite{rombach2022high} for each subject, in which the generating prompts are obtained from Claude-3-Opus \cite{claude3}.
Then, to guarantee the co-occurrence of multiple subjects, similarly, we concatenate the class images of multiple subjects.
The concatenated class image is then fed into the U-Net $\theta$ for training with the text prompt ``a cat and a dog'' (taking cat and dog as an example).
To be note that these images are also conducted background removal.
As there is no learnable text tokens during prior preservation training, the attention loss is not applied. We only apply the reconstruction loss for supervision as the following:
\begin{equation}
\mathcal{L}_{recon}^{pr}=\mathbb{E}_{\epsilon^{\prime}, \boldsymbol{z}^{\prime}, c^{\prime}, t^{\prime}}\left[\left\|\epsilon^{\prime}-\epsilon_\theta^{\prime}\left(\boldsymbol{z}_{t^{\prime}}^{\prime}, \mathcal{T}(c^{\prime}), t^{\prime}\right)\right\|_2^2\right],
\label{equ:t2v_prior}
\end{equation}
where $\boldsymbol{z}^\prime$ is the latent code of the input concatenated class image, $\epsilon_\theta^\prime$ is the noise prediction from the model $\theta$, $\epsilon^\prime$ is the randomly sampled Gaussian noise, $c^{\prime}$ is the text condition for the concatenated class image, and $t^\prime$ is the sampling timestep.
To this end, we train our CustomVideo via an end-to-end manner, with the following overall training objective:
\begin{equation}
    \mathcal{L}=\mathcal{L}_{recon} + \alpha\cdot\mathcal{L}_{attn}^{pos\_neg} + \beta\cdot\mathcal{L}_{recon}^{pr}.
    \label{equ:overall}
\end{equation}
where $\alpha$ and $\beta$ are two hyper-parameters to control the weight of attention control and prior preservation, respectively.
Our proposed co-occurrence and attention control mechanism is exclusively applied during model training to assist the model and learnable text tokens in disentangling the identities of multiple subjects. During the training stage, images of different subjects are carefully concatenated without overlap, enabling the attention control mechanism to accurately locate the regions of each subject and learn their corresponding features.



\paragraph{Inference} 
During inference, CustomVideo only requires a specific text prompt with corresponding learned word token integrated to generate a required video.
To be note that ground truth masks of subjects are not required during inference. 
Thanks to the thorough optimization of the model and learnable text tokens during training, even challenging scenarios with overlapping subjects—such as the "car and barn" example—can be effectively managed (see Fig. \ref{fig:teaser}).
Although the model is fine-tuned with single-frame videos extended from subject images, our proposed CustomVideo can generate videos with high diversity while not losing the the fidelity of subjects and motion smoothness (see Fig. \ref{fig:teaser}).

\section{Experiment}

\begin{figure}[t]
  \centering
   \includegraphics[width=\linewidth]{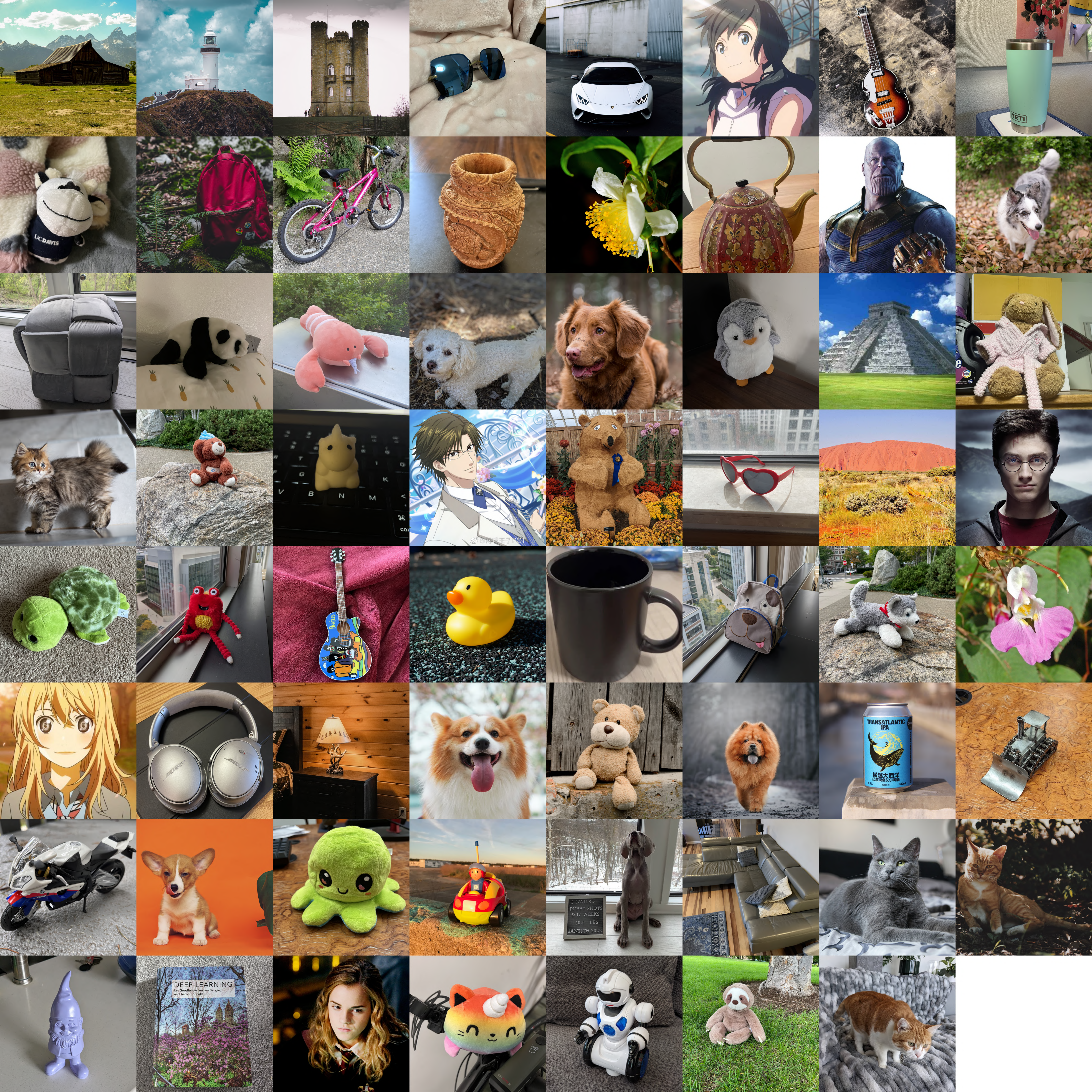}
   \vspace{-5mm}
   \caption{{\bf The overview of our CustomStudio dataset.} Samples in our CustomStudio dataset cover a wide range of 13 object categories, including anime character, decorate item, furniture, instrument, person, pet, plant, plush, scene, thing, toy, transport, and wearable item.}
   \label{fig:customstudio}
\end{figure}

\begin{table}[t]
  \centering
      \caption{\textbf{Subject pairs} constructed upon categories of the subjects in our CustomStudio dataset.}
      \vspace{-1mm}
  \label{tab:subject_pair}%
\scalebox{0.8}{

\begin{tabular}{l|l}
\toprule
\multirow{5}[2]{*}{Two-subject Pair} & [decorate item, plant], [toy, toy], [pet, pet], [furniture, thing] \\
      & [decorate item, furniture], [instrument, person], [person, person] \\
      & [anime character, anime character], [person, wearable item] \\
      & [plush, plush], [person, transport], [instrument, anime character] \\
      & [person, scene], [scene, transport], [toy, transport] \\
\midrule
\multirow{5}[2]{*}{Three-subject Pair} & [anime character, anime character, scene], [toy, toy, toy] \\
      & [anime character, instrument, scene], [plush, plush, plush] \\
      & [person, instrument, scene], [person, person, scene] \\
      & [pet, wearable item, scene], [toy, transport, scene] \\
      & [pet, pet, scene], [decorate item, furniture, furniture] \\
\bottomrule
\end{tabular}%

}

\end{table}%

\begin{table}[t]
  \centering

  \caption{\textbf{Prompt templates} used for generating videos during inference.}
  \label{tab:prompt_templates}
      \vspace{-1mm}
  \scalebox{0.74}{

\begin{tabular}{l}
\toprule
\textbf{Prompt Template for Two-subject Pair } \\
\midrule
\midrule
a [c1] and a [c2] sitting on an antique table \\
a [c1] and a [c2] sitting on beach with a view of seashore \\
a [c2] and a [c1] side by side on a mountaintop, overlooking a sunrise \\
a [c1] and a [c2] on a surfboard together in the middle of a clear blue ocean \\
a [c1] playing with [c2] \\
a [c1] playing with a robot toy [c2] \\
a [c2] playing with a robot toy [c1] \\
a plush toy replica of a [c1] and a [c2] sitting beside it \\
a [c1] and a [c2] walking in the Times Square \\
a [c1] and a [c2] walking on the Great Wall \\
\midrule
\textbf{Prompt Template for Three-subject Pair } \\
\midrule
\midrule
a [c1] wearing a [c2], with a [c3] in the background \\
a [c1] wearing a [c2] and giving a speech, with a [c3] in the background \\
a wide shot of a [c1] wearing a [c2] with boston city in background, with a [c3] in the background \\
a long shot of a [c1] walking their golden retriever, wearing a [c2], with a [c3] in the background \\
a [c1] sitting on the sidewalk wearing a [c2], with a [c3] in the background \\
a [c1] standing at a graffiti wall, showcasing a new [c2], with a [c3] in the background \\
close shot of a [c1] walking on a ramp wearing a [c2], with a [c3] in the background \\
a [c1] stepping out of a taxi in a [c2], with a [c3] in the background \\
close shot of a [c1] walking in rain wearing a [c2], with a [c3] in the background \\
long shot of a [c1] sitting on the edge of a roof wearing [c2], with a [c3] in the background \\
\bottomrule
\end{tabular}%

    }

\end{table}%

\begin{table}[t]
  \centering

    \caption{\textbf{Class prompts} for generating class images used in class-specific prior preservation. In our experiments, we generate one class image with one independent text prompt. Here we show 20 examples for 20 class images.}
    \label{tab:class_prompts}%
      \vspace{-1mm}
  
\scalebox{0.8}{

    \begin{tabular}{l}
    \toprule
    A fluffy white Persian cat with blue eyes, sitting on a plush red cushion \\
    A playful orange tabby kitten chasing a ball of yarn across a hardwood floor \\
    A sleek black cat with green eyes, perched on a windowsill, looking out at a rainy day \\
    A cute Siamese cat with a pink collar, napping in a wicker basket \\
    A majestic Maine Coon cat with a bushy tail, walking through a field of wildflowers \\
    A curious Sphynx cat with wrinkled skin, exploring a cardboard box \\
    A lazy gray British Shorthair cat, curled up on a soft blanket, basking in the sunlight \\
    A mischievous Calico cat, peeking out from behind a potted plant \\
    A regal Russian Blue cat with silver fur, sitting on a velvet armchair \\
    A friendly Ragdoll cat with bright blue eyes, being petted by a gentle hand \\
    A playful Bengal cat with spots, pouncing on a feather wand toy \\
    A curious Abyssinian cat with large ears, investigating a paper bag \\
    A fluffy Himalayan cat with a flat face, lounging on a fuzzy rug \\
    A sleepy Birman cat with white paws, dozing off on a cozy bed \\
    A graceful Siberian cat with long fur, walking along a wooden fence \\
    A cute Munchkin cat with short legs, playing with a crinkly ball \\
    A curious Norwegian Forest cat with a thick coat, climbing a cat tree \\
    A playful Scottish Fold cat with folded ears, batting at a dangling string \\
    A majestic Savannah cat with a tall, slender build, surveying its surroundings \\
    A friendly Tonkinese cat with a unique coat pattern, rubbing against its owner's leg \\
    \bottomrule
    \end{tabular}%

    }

\end{table}%

\subsection{Experimental Setup}
\label{sec:experimental_setup}

\paragraph{Dataset} 

We collect a dataset CustomStudio for multi-subject driven video generation.
This dataset is composed of 63 individual objects and 68 meaningful pairs.
These objects cover a wide range of 13 diverse categories, including anime character, decorate item, furniture, instrument, person, pet, plant, plush, scene, thing, toy, transport, and wearable item.
For each subject, 4-21 images are provided as the reference images.
The images are adopted from DreamBooth \cite{ruiz2023dreambooth}, CustomDiffusion \cite{kumari2023multi} and Mix-of-Show \cite{gu2023mix}.
An overview of the samples in CustomStudio dataset is shown in Fig.\ref{fig:customstudio}.
A meaningful pair is composed of different subjects may appear in a video together in practice. For example, `cat' and `dog' can be a meaningful pair, while `book' and `barn' can not arise in the most cases. 
We set up 25 types of subject pairs, which are constructed based on the corresponding categories, as shown in Table \ref{tab:subject_pair}.
Each pair of subjects has 10 different text prompts for evaluation, which are designed with different contexts, actions, and so on, as shown in Table \ref{tab:prompt_templates}.
The text prompts for generating are obtained from Claude-3-Opus \cite{claude3}. 
We show 20 examples in Table \ref{tab:class_prompts}.

\begin{figure*}[htbp]
  \centering
   \includegraphics[width=\linewidth]{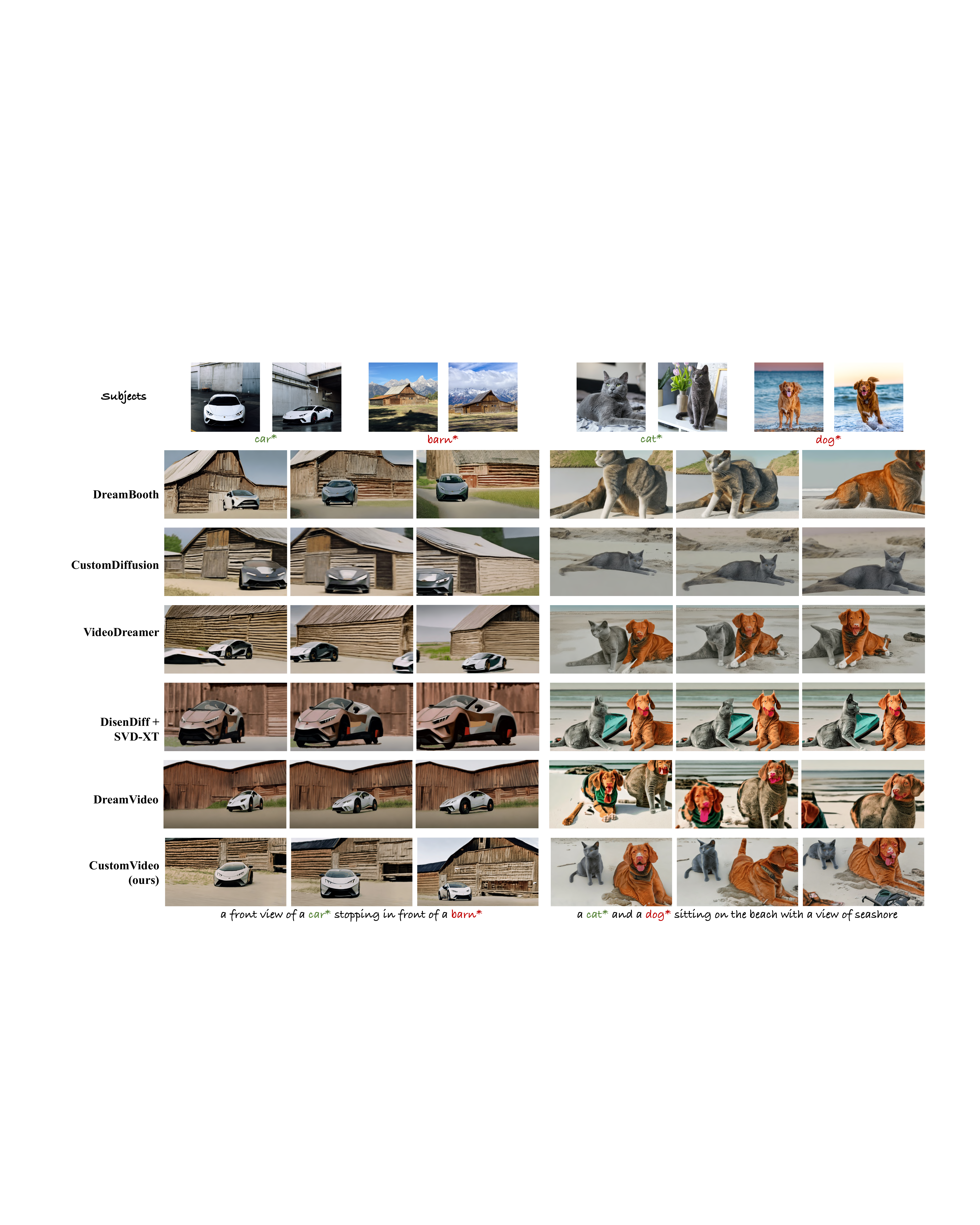}
   \vspace{-5mm}
   \caption{{\bf Qualitative results of our CustomVideo with comparison to SOTA methods}, including DreamBooth, CustomDiffusion, VideoDreamer, DisenDiff + SVD-XT, and DreamVideo. The first line indicates the given subjects, while each line indicates the frames generated by the corresponding method. The bottom line is the text prompt for inference. We can observe that our CustomVideo can generate videos with better fidelity of subjects compared with previous SOTA methods.}
   \label{fig:sota_comparison}
\end{figure*}


\paragraph{Implementation Details} 
We train CustomVideo for 500 steps with AdamW \cite{loshchilov2017decoupled} optimizer (batch size 2, learning rate 4e-5, and weight decay 1e-2).
For class-specific prior preservation, we generate 200 class images with Stable Diffusion v2.1 \cite{rombach2022high} for each subject, in which the generating prompts are obtained from Claude-3-Opus \cite{claude3}.
Note that the corresponding class images are also concatenated during training phase.
The negative value $\eta$ in the mask $\boldsymbol{\mathcal{M}}$ is set as -1e-8.
The weight parameters $\alpha$ and $\beta$ in Eq. \eqref{equ:overall} are set as 0.2 and 1.0, respectively.
During inference, we perform 50 steps denoising with DPM-Solver \cite{lu2022dpm} sampler and classifier-free guidance \cite{ho2022classifier}.
The resolution of generated 24-frame videos are $576\times320$ with 8 fps.
Interestingly, we find that the trained weights and word tokens learned from low resolution T2V model can be directly loaded for high resolution zeroscope T2V model \cite{zeroscopexl} to generate personalized videos with $1024\times576$ resolution, in which no additional training computation cost is required.
The weights of key and value in all of the cross-attention layers (including the temporal and spatial layers) of main blocks within U-Net are fine-tuned.
To be note that an individual temporal transformer apart from main blocks within U-Net is responsible for modeling the motion in video generation. This temporal component remains fixed throughout our training process.
And we only focus on fine-tuning a small amount of key and value parameters. This targeted optimization allows the model to learn the identities of the provided multiple subjects while leaving the motion generation capability of the base model entirely intact.
The sampling step for high resolution video is 30.
Our CustomVideo is implemented based on PyTorch \cite{paszke2019pytorch} and Diffusers \cite{diffusers2022}.
All of the training process is under fp16 mixed precision with accelerate package \cite{accelerate2022sylvain}.
We use data augmentations during training, including randomly horizontal flip, randomly crop and resize, together with corresponding prompt change (`very small' or `close up' appends before the prompt).
The training phase takes about 8 minutes for a subject pair on 1 RTX 3090 GPU.
Meanwhile, it takes about 1 minute and 2 minutes to generate a low and high resolution video on 1 RTX 3090 GPU, respectively.

\paragraph{Prompt Templates}
The prompt templates used for generating videos during inference are shown in Table \ref{tab:prompt_templates}.
Taking ``a [c1] and a [c2] walking in the Times Square'' as an example, ``[c1]'' and ``[c2]'' are two positions for integrating the paired class names and learnable text tokens.
If we want to generate a video with ``cat'' and ``dog'', we can integrate these names into the above prompt template with the corresponding learned text tokens ``$\textless$new1$\textgreater$'' and ``$\textless$new2$\textgreater$''.
To this end, the prompt for inference comes to ``a $\textless$new1$\textgreater$ cat and a $\textless$new2$\textgreater$ dog walking in the Times Square'' or ``a $\textless$new1$\textgreater$ dog and a $\textless$new2$\textgreater$ cat walking in the Times Square''.
Both two prompts works for customized generation.

\paragraph{Comparison Methods} 
Except for most relevant VideoDreamer \cite{chen2023videodreamer}, we also consider adapting previous SOTA image-based multi-subject driven methods to a video scenario for comparison, including DreamBooth \cite{ruiz2023dreambooth} and CustomDiffusion \cite{kumari2023multi}.
Also, we compare our method with single-subject driven video personalization method DreamVideo \cite{wei2024dreamvideo}.
Moreover, we consider an alternative way for multi-subject T2V customization, that is, the combination of multi-concept T2I and I2V diffusion models.
For this case, we utilize the latest multi-concept driven T2I method DisenDiff \cite{zhang2024attention} for customization and I2V method SVD-XT \cite{blattmann2023stable} for animating the generated image.
We adopt the follow settings for the comparison methods in our experiments:
\begin{itemize}
\item {\bf DreamBooth} \cite{ruiz2023dreambooth}: 
For efficient fine-tuning, we utilize LoRA \cite{hu2022lora} to adapt the U-Net under DreamBooth \cite{ruiz2023dreambooth}.
The rank of LoRA module is set as 4.
We use AdamW \cite{loshchilov2017decoupled} optimizer, with learning rate 4e-4, weight decay 1e-2, batch size 2, and training steps 1000.
\item {\bf CustomDiffusion} \cite{kumari2023multi}:
The weights of query and value in all of the cross-attention layers of U-Net are fine-tuned in CustomDiffusion \cite{kumari2023multi}.
We use AdamW \cite{loshchilov2017decoupled} optimizer, with learning rate 4e-5, weight decay 1e-2, batch size 2, and training steps 500.
\item {\bf VideoDreamer} \cite{chen2023videodreamer}: 
LoRA \cite{hu2022lora} is utilized for fine-tuning the U-Net and text encoder in VideoDreamer \cite{chen2023videodreamer}.
The AdamW \cite{loshchilov2017decoupled} optimizer is used  with learning rate 5e-5, weight decay 1e-2, batch size 2, and training steps 500.
\item {\bf DisenDiff} \cite{zhang2024attention} + {\bf SVD-XT} : 
DisenDiff is trained for 250 steps with AdamW \cite{loshchilov2017decoupled} optimizer (learning rate 4e-5, weight decay 1e-2, batch size 2).
The sampling step for SVD-XT is 30.
\item {\bf DreamVideo} \cite{wei2024dreamvideo}:
We train the learnable textual token and identity adapter for 3000 and 800 steps, respectively, with learning rate 1e-4 and 1e-5. We use AdamW \cite{loshchilov2017decoupled} optimizer with a batch size of 2.
\end{itemize}



\begin{figure*}[htbp]
  \centering
   \includegraphics[width=\linewidth]{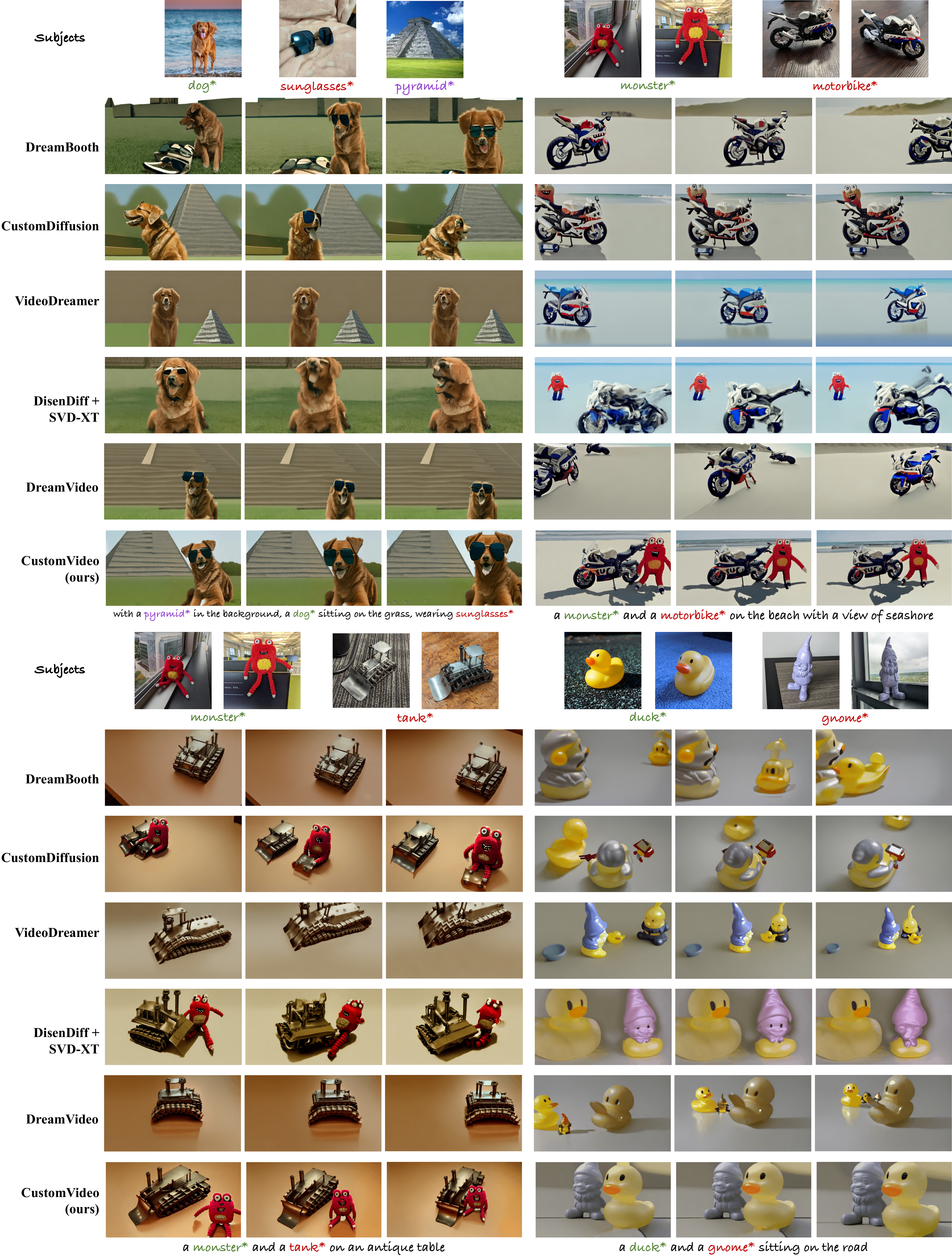}
   \caption{
   \textbf{Additional comparison results} of our CustomVideo with previous SOTA methods.
   }
   \label{fig:sota_comparison_additional}
\end{figure*}

\begin{figure*}[htbp]
  \centering
   \includegraphics[width=\linewidth]{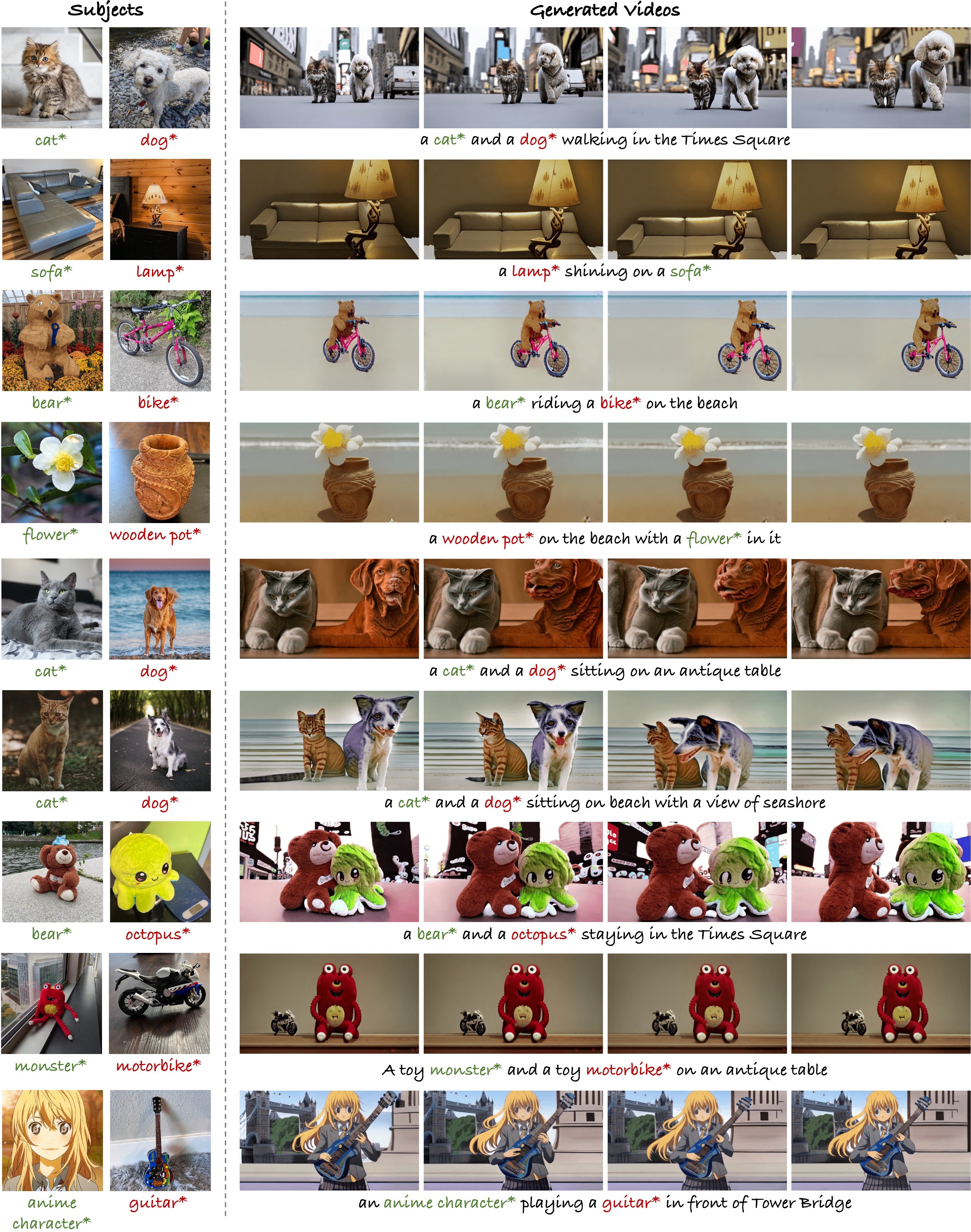}
   \caption{
   \textbf{Qualitative results} of generated videos from our CustomVideo. It is clear that our method can generate videos with higher fidelity and better preserve the identities of the reference subjects.
   }
   \label{fig:additional_results}
\end{figure*}

\begin{table}[t]
  \centering

    \caption{\textbf{Quantitative  results of our CustomVideo with comparison to SOTA methods}, including DreamBooth, CustomDiffusion, VideoDreamer, DisenDiff + SVD-XT, and DreamVideo. Our proposed CustomVideo consistently outperforms previous SOTA methods for all 4 evaluation metrics.}
  \label{tab:sota_comparison}%
  \vspace{-1mm}
  
\setlength\tabcolsep{2pt}
  \scalebox{0.9}{


\begin{tabular}{lccccc}
\toprule
Method & \multicolumn{1}{l}{Venue} & CLIP-T $\uparrow$ & CLIP-I $\uparrow$ & DINO-I $\uparrow$ & T. Cons. $\uparrow$ \\
\midrule
\midrule
DreamBooth \cite{ruiz2023dreambooth} & \multicolumn{1}{l}{CVPR'23} & 0.6451  & 0.6079  & 0.3109  & 0.7084  \\
CustomDiffusion \cite{kumari2023multi} & \multicolumn{1}{l}{CVPR'23} & 0.6524  & 0.6206  & 0.3164  & 0.7342  \\
VideoDreamer \cite{chen2023videodreamer} & \multicolumn{1}{l}{arXiv'23} & 0.6638  & 0.6297  & 0.3479  & 0.7267  \\
DisenDiff \cite{zhang2024attention} + SVD \cite{blattmann2023stable} & \multicolumn{1}{l}{CVPR'24} & 0.6592  & 0.6179  & 0.3318  & 0.7095  \\
DreamVideo \cite{wei2024dreamvideo} & \multicolumn{1}{l}{CVPR'24} & 0.6646  & 0.6128  & 0.3228  & 0.7498  \\
\midrule
\rowcolor[rgb]{ .851,  .906,  .992} CustomVideo & \multirow{2}[2]{*}{} & \textbf{0.7075 } & \textbf{0.6863 } & \textbf{0.3983 } & \textbf{0.7960 } \\
\rowcolor[rgb]{ .851,  .906,  .992} (ours vs. DreamVideo \cite{wei2024dreamvideo}) &       & \textcolor[rgb]{ .753,  0,  0}{+6.46\%} & \textcolor[rgb]{ .753,  0,  0}{+11.99\%} & \textcolor[rgb]{ .753,  0,  0}{+23.39\%} & \textcolor[rgb]{ .753,  0,  0}{+6.16\%} \\
\bottomrule
\end{tabular}%

    }

\end{table}%

\paragraph{Evaluation Metrics} 

Following previous works \cite{wei2024dreamvideo,chen2023videodreamer}, we quantitatively evaluate our CustomVideo with the following 4 metrics: 1) \emph{CLIP Textual Alignment} computes the average cosine similarity between the generated frames and text prompt with CLIP \cite{radford2021learning} ViT-B/32 \cite{dosovitskiy2021an} image and text models; 2) \emph{CLIP Image Alignment} calculates the average cosine similarity between the generated frames and subject images with CLIP ViT-B/32 image model; 3) \emph{DINO Image Alignment} measures the average visual similarity between generated frames and reference images with DINO \cite{caron2021emerging} ViT-S/16 model; 4) \emph{Temporal Consistency} \cite{esser2023structure} evaluates the average cosine similarity of all consecutive frame pairs in the generated videos.

\subsection{Main Results}



\paragraph{Qualitative Results} 
We present the qualitative comparison results in Fig.\ref{fig:sota_comparison}. 
From these results, we observe that our CustomVideo approach effectively ensures the co-occurrence of multiple subjects and successfully disentangles different subjects. 
However, both DreamBooth and CustomDiffusion fail to capture the structural color information of the car provided in our experiments. 
The dominant black color from the car window obscures the entire car, resulting in low fidelity. 
Moreover, the generated frames from VideoDreamer lack consistency, as some frames depict two cars while others only show one car. 
Additionally, VideoDreamer fails to capture accurate color information, such as the color of the car door. 
The alternative approach, DisenDiff + SVD-XT can not neither retain the identities of the given subjects nor generate videos with rich motions.
Without specific design for tackling multiple subjects, DreamVideo can not distinguish the similar pair `cat' and `dog'.
In contrast, our CustomVideo method excels in handling such challenging scenes and foreground subject scenarios, effectively capturing the intricate structural details of the provided car for video generation. 
Similarly, when considering the case of `cat' and `dog', our approach also demonstrates superior capability in generating high-quality videos.
More results can be found in Fig.\ref{fig:sota_comparison_additional} and Fig.\ref{fig:additional_results}.

\paragraph{Quantitative Results} 
We conduct quantitative experiments on our collected dataset. 
To ensure a thorough analysis, we generate videos using 10 individual prompts and 4 random seeds for each pair of subjects, resulting in a total of 2,720 generated videos for each method. 
We then evaluate the quality of the generated videos using four metrics, and the results are presented in Table \ref{tab:sota_comparison}. 
The table clearly demonstrates that our proposed method is capable of generating videos that are better aligned with the given subjects, outperforming the most recent video diffusion model, DreamVideo, by 11.99\% and 23.39\% in terms of \emph{CLIP Image Alignment} and \emph{DINO Image Alignment}, respectively.
Meanwhile, the videos generated from our approach achieve better alignment with the desired text prompts, exceeding DreamVideo by 6.46\% \emph{CLIP Textual Alignment}.
These improvements can be attributed to our specially designed co-occurrence and attention control mechanisms, which effectively disentangle and preserve the fidelity of the subjects. 
Furthermore, our CustomVideo generates videos with significantly higher temporal consistency compared to SOTA methods, as indicated in Table \ref{tab:sota_comparison}. 
For instance, CustomVideo surpasses DreamVideo by 6.16\% in terms of \emph{Temporal Consistency}.
Moreover, CustomVideo outperforms DisenDiff+SVD-XT by +11.07\% in CLIP Image Alignment and +12.19\% in Temporal Consistency, demonstrating the effectiveness and robustness of our method in preserving the identity and modeling the motion.

\begin{figure}
\centering
\includegraphics[width=\linewidth]{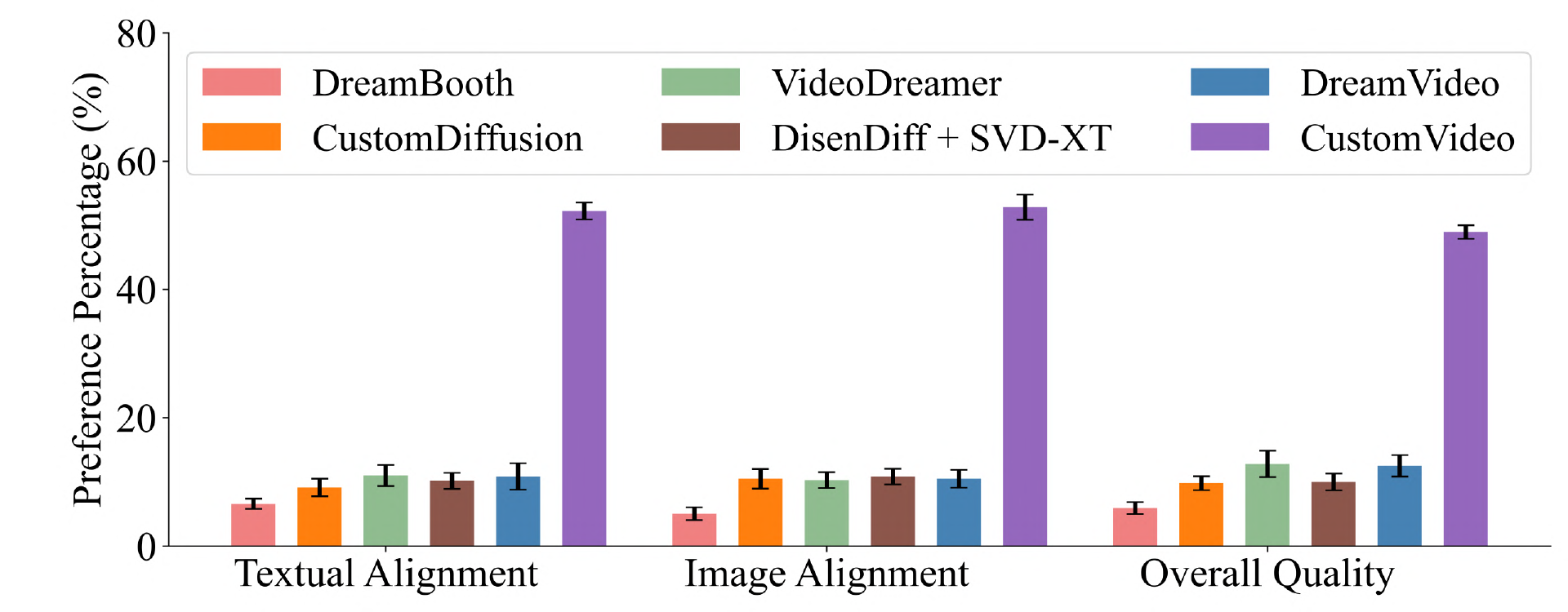}
\vspace{-5mm}
\caption{{\bf User study}. Our CustomVideo achieves the best human preference compared with 5 SOTA comparison methods in terms of \emph{Textual Alignment}, \emph{Image Alignment}, and \emph{Overall Quality}.}
\label{fig:user_study}
\end{figure}


\begin{figure}[t]
  \centering
   \includegraphics[width=\linewidth]{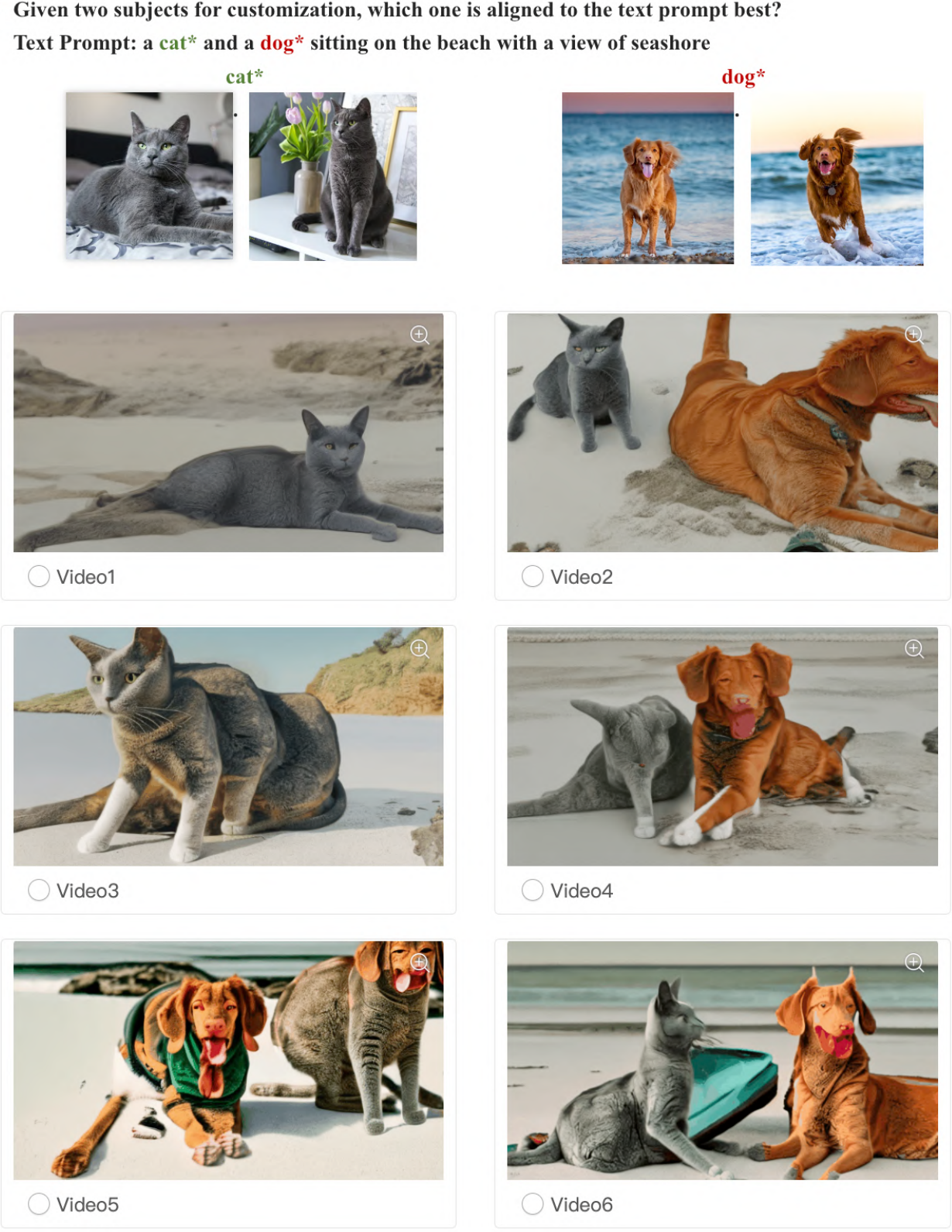}
   \vspace{-5mm}
   \caption{
   \textbf{Human interface} used in the user study. The generated videos of 5 comparison approaches and our CustomVideo are randomly ordered and anonymous for fair comparison.
   }
   \label{fig:human_interface}
\end{figure}

\paragraph{Human Preference Study} 
To further validate our method, we conduct human evaluations on our CustomVideo with comparison to 5 SOTA methods.
In this study, we collect 1500 answers from 25 independent human raters with the following questions: 1) which one is aligned to the text prompt best? 2) which one is aligned to the subject images best? 3) which one has the best overall quality?
The results are shown in Fig.\ref{fig:user_study}.
Our CustomVideo is found to be the most preferred option based on human evaluations in all three dimensions.
The human annotators are all researchers with experience in artificial intelligence. The interface for human evaluation is shown in Fig.\ref{fig:human_interface}.




\subsection{Ablation Studies} 
We conduct ablation analysis of our method from 3 aspects: 1) the effect of each component in our method; 2) the behavior of attention control; 3) the effect of cross-attention maps from different layers; 4) the effect of hyper-parameters in attention control.



\paragraph{Component Analysis}

\begin{figure}
\centering
\includegraphics[width=\linewidth]{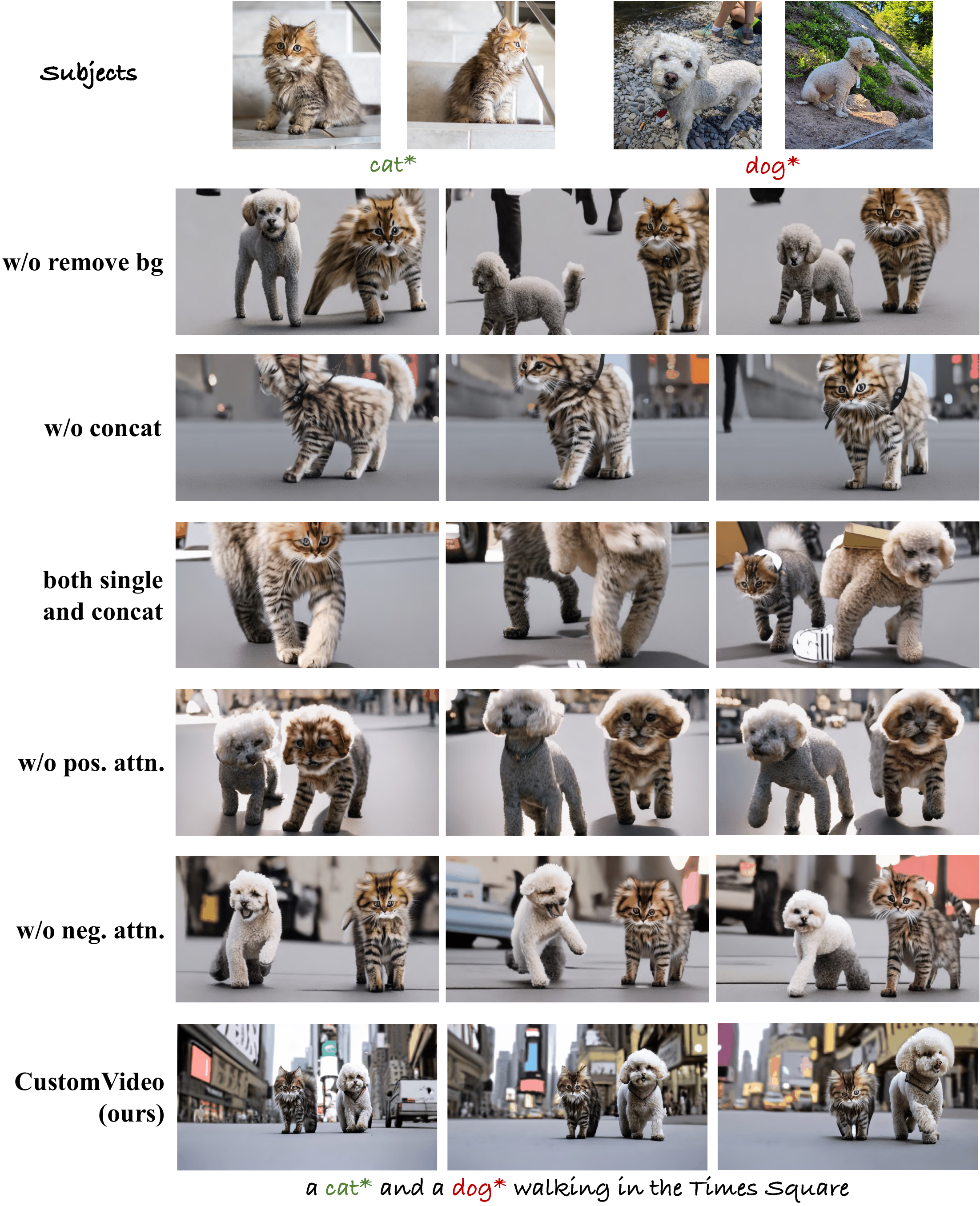}
\vspace{-6mm}
\caption{{\bf Qualitative results for component analysis} of our proposed CustomVideo. We find that ensuring concatenating subjects during training is effective for guaranteeing the co-occurrence in the generated video. Moreover, our attention mechanism could disentangle different subjects.}
\label{fig:component_analysis}
\end{figure}

We conduct a thorough analysis of each component in our CustomVideo, presenting both qualitative and quantitative results in Fig.\ref{fig:component_analysis} and Table \ref{tab:component_analysis}. 
One crucial observation is the significance of ensuring co-occurrence during the fine-tuning process. 
We notice a significant drop in performance when multiple subjects are not concatenated into a single image (line `w/o concat' in Table \ref{tab:component_analysis}). 
The absence of subject concatenation leads to the domination of one single subject, which is deemed unacceptable (line `w/o concat' in Fig.\ref{fig:component_analysis}). 
Moreover, we conduct a study on fine-tuning T2V model with both single and concatenated subjects to examine if the single subject could aid in learning corresponding characteristics. 
Surprisingly, the results reveal that adding the single subjects to the training process proved detrimental, resulting in inconsistent generation during inference (line `both single and concat' in Fig.\ref{fig:component_analysis}).

\begin{figure*}[htbp]
  \centering
   \includegraphics[width=\linewidth]{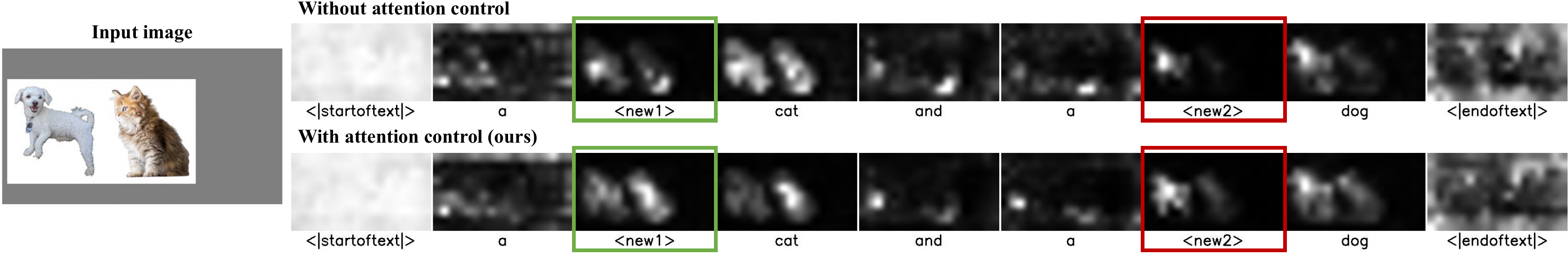}
   \vspace{-5mm}
   \caption{{\bf Comparison results of cross-attention maps}. We observe that the learnable word tokens better match the corresponding correct areas of the subjects with our proposed attention control.
   }
   \label{fig:attention_visualization}
\end{figure*}

\begin{table}[t]
    \centering
    
\caption{\textbf{Quantitative results for component analysis} of our proposed CustomVideo. We observe a significant performance drop when removing subjects concatenation or background removal. Moreover, training the model with both single and concatenated subjects will produce unstable results, leading to performance degradation. In practice, we suggest simultaneously using positive and negative attention mechanisms to obtain the best results.} 
\label{tab:component_analysis}
\vspace{-1mm}

    \setlength\tabcolsep{3pt}
  \scalebox{0.9}{
\begin{tabular}{lcccc}
\toprule
Method & CLIP-T$\uparrow$ & CLIP-I$\uparrow$ & DINO-I$\uparrow$ & T. Cons.$\uparrow$ \\
\midrule
\midrule
w/o remove bg & 0.6682  & 0.6752  & 0.3636  & 0.7609  \\
w/o concat & 0.6115  & 0.6611  & 0.3386  & 0.7034  \\
both single and concat & 0.6248  & 0.6096  & 0.3054  & 0.7604  \\
w/o pos. attn. & 0.6851  & 0.6158  & 0.3215  & 0.7213  \\
w/o neg. attn. & 0.6961  & 0.6716  & 0.3613  & 0.7886  \\
\midrule
\rowcolor[rgb]{ .851,  .906,  .992} {\bf CustomVideo (ours)} & \textbf{0.7075} & \textbf{0.6863} & \textbf{0.3983} & \textbf{0.7960} \\
\bottomrule
\end{tabular}%
}

\end{table}

\begin{figure*}[t]
  \centering
   \includegraphics[width=\linewidth]{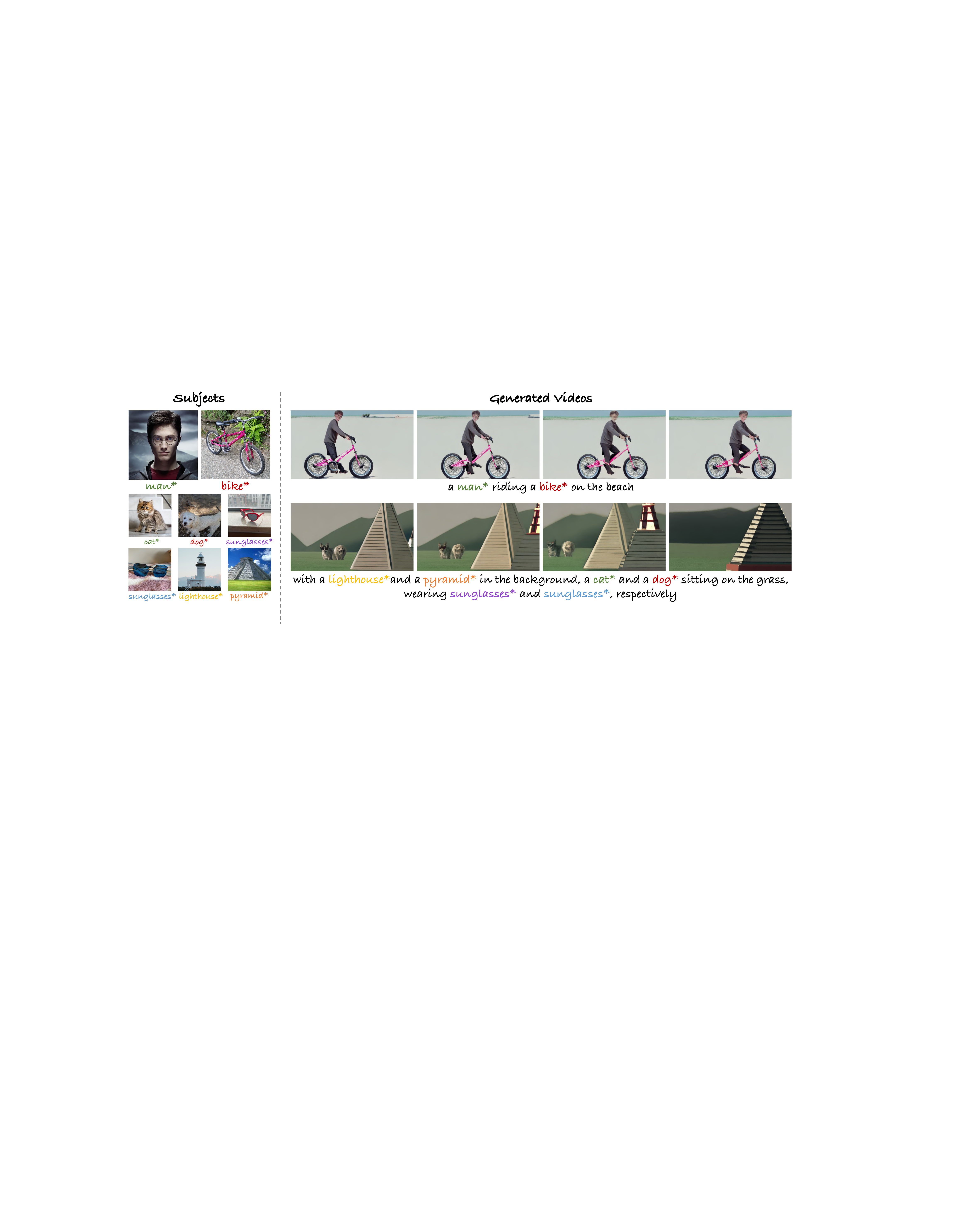}
   \vspace{-6mm}
   \caption{{\bf Failure cases of our CustomVideo.} Our CustomVideo fails to generate vivid facial contents with a global view, such as `man' and `bike' (1st line). Also, our approach can not tackle too many subjects, \eg, customizing T2V generation with 6 different subjects (2nd line).}
   \label{fig:failure_cases}
\end{figure*}


Our attention control mechanism plays a crucial role in preserving the identities of subjects. 
The positive attention guidance in particular improves the \emph{CLIP Image Alignment} metric by 11.45\%, as demonstrated in Table \ref{tab:component_analysis}. 
The negative attention guidance also proved beneficial in promoting better image alignment. 
Moreover, the attention control mechanism significantly enhances the temporal consistency of the generated videos, a crucial factor in video generation. 
The videos generated without positive and negative attention guidance clearly showcase the positive impact of both components (lines `w/o pos. attn.' and `w/o neg. attn.' in Fig.\ref{fig:component_analysis}). 
The positive guidance specifically preserves the unique characteristics of a subject, such as distinguishing between a cat and a dog, while the negative guidance weakens the influence of other subjects on a specific subject. 
By incorporating both positive and negative guidance, our CustomVideo excels in generating videos with high subject fidelity and remarkable temporal consistency.
Furthermore, we observe that removing the background from the given subject images significantly improves the generation of high-quality videos (line `w/o remove bg' in Fig.\ref{fig:component_analysis}).
By eliminating the background, the T2V model can focus solely on learning the characteristics of the given subjects.

\paragraph{Comparison of Cross-attention Maps}
\label{par:cross_attention_maps}
We investigate the cross-attention maps during training, and study the behavior of our attention control mechanism. 
The visualization results are shown in Fig.\ref{fig:attention_visualization}.
Without attention control, it can be observed that the learnable word token `$\textless$new1$\textgreater$' for cat is aligned to the region of dog, which is unacceptable.
In contrast, equipped with attention control mechanism,  the learnable word token `$\textless$new1$\textgreater$' and `$\textless$new2$\textgreater$' are better aligned to the correct areas of cat and dog, respectively.
Such results strongly demonstrate the effectiveness of our approach.

\paragraph{Layers of Cross-attention Maps}
We investigates the effect of how we extract the cross-attention maps from the cross-attention layers.
There are four levels of cross-attention layers in the U-Net, in which the sizes cross-attention maps are different.
Taking the resolution $576\times320$ of our generated low resolution video as an example, the sizes of cross-attention maps are $72\times40$ ($\mathcal{\ell}_1$), $36\times20$ ($\mathcal{\ell}_2$), $18\times10$ ($\mathcal{\ell}_3$), and $9\times5$ ($\mathcal{\ell}_4$).
Here, we study the effect of different levels of cross-attention maps, the results are shown in Table \ref{tab:cross_attention_level}.
To tackle the size mismatch between the cross-attention maps and ground truth mask, we systematically downsample the ground truth mask to precisely match the spatial dimensions of different cross-attention maps. Specifically, when utilizing cross-attention maps from both \(\mathcal{\ell}_1\) (\(72 \times 40\)) and \(\mathcal{\ell}_2\) (\(36 \times 20\)), the ground truth mask is resized to the respective size levels of \(\mathcal{\ell}_1\) and \(\mathcal{\ell}_2\) simultaneously. Attention control is then performed at both \(\mathcal{\ell}_1\) and \(\mathcal{\ell}_2\) levels to evaluate their combined impact on subject disentanglement and feature learning.
We observe a significant performance drop with too large ($\mathcal{\ell}_1$) or small ($\mathcal{\ell}_4$) cross-attention maps.
Large cross-attention maps may, on one hand, lead to the loss of subtle subject features, while on the other hand, small cross-attention maps can result in inaccurate optimization between the maps and the downsampled ground truth masks.
Regarding the cross-attention maps with middle size from $\mathcal{\ell}_2$ and $\mathcal{\ell}_3$, $\mathcal{\ell}_3$ works better.
Thus, in our experiments, we extract cross-attention maps from $\mathcal{\ell}_3$ for attention control.

\begin{table}[t]
  \centering

  \caption{\textbf{Quantitative results of cross-attention levels.} We find that only using cross-attention map from $\ell_3$ works best and adopt this setting by default.}
  \label{tab:cross_attention_level}
      \vspace{-1mm}
  
\setlength\tabcolsep{3pt}
  \scalebox{0.9}{

\begin{tabular}{cccc|cccc}
\toprule
$\mathcal{\ell}_1$ & $\mathcal{\ell}_2$ & $\mathcal{\ell}_3$ & $\mathcal{\ell}_4$ & CLIP-T$\uparrow$ & CLIP-I$\uparrow$ & DINO-I$\uparrow$ & T. Cons.$\uparrow$ \\
\midrule
\midrule
\Checkmark & \XSolidBrush & \XSolidBrush & \XSolidBrush & 0.6111  & 0.6335  & 0.3964  & 0.7315  \\
\XSolidBrush & \Checkmark & \XSolidBrush & \XSolidBrush & 0.6080  & 0.6641  & 0.3055  & 0.7242  \\
\rowcolor[rgb]{ .851,  .906,  .992} \XSolidBrush & \XSolidBrush & \Checkmark & \XSolidBrush & \textbf{0.7075 } & \textbf{0.6863 } & \textbf{0.3983 } & \textbf{0.7960 } \\
\XSolidBrush & \XSolidBrush & \XSolidBrush & \Checkmark & 0.6134  & 0.6420  & 0.3215  & 0.7793  \\
\Checkmark & \Checkmark & \XSolidBrush & \XSolidBrush & 0.6234  & 0.6375  & 0.3497  & 0.7443  \\
\XSolidBrush & \XSolidBrush & \Checkmark & \Checkmark & 0.7047  & 0.6007  & 0.3642  & 0.7380  \\
\XSolidBrush & \Checkmark & \Checkmark & \XSolidBrush & 0.6730  & 0.6364  & 0.3977  & 0.7433  \\
\Checkmark & \XSolidBrush & \XSolidBrush & \Checkmark & 0.6508  & 0.6360  & 0.3055  & 0.7469  \\
\Checkmark & \Checkmark & \Checkmark & \Checkmark & 0.6432  & 0.6738  & 0.3594  & 0.6975  \\
\bottomrule
\end{tabular}%

   }

\end{table}%

\paragraph{Effect of Hyper-parameters in Attention Control}
We also conduct investigations into the effects of two important parameters: the weight of attention loss ($\alpha$) and the negative value ($\eta$) used in the guidance mask ($\boldsymbol{\mathcal{M}}$). 
The quantitative results of these investigations are presented in Table \ref{tab:attention_weight} and Table \ref{tab:negative_value}.
We observe that the best performance is achieved when the weight $\alpha$ is set to 0.2. 
This value results in optimal alignment between the generated videos and the given subjects, indicating the importance of appropriately balancing the attention guidance during training.
For the negative value in the guidance mask, our findings reveal that even a slight negative value is sufficient to enhance the quality of T2V generation. 
This implies that incorporating negative attention guidance can effectively suppress the influence of other subjects on a specific subject, improving the generation quality.

\subsection{Limitations}
In Fig.\ref{fig:failure_cases}, we present some failure cases encountered during our experiments. 
Since our method relies on the ability of the base model, it would fail if the base model could not generate, such as small faces in a global view. 
Currently, Our approach can not tackle too many subjects, in which the spatial positions of different objects in the video conflict with each other.
To address such case, we can control the positions of different subjects via conditioning on the pre-defined spatial bounding boxes during inference sampling.


\begin{table}[t]
    \centering
    
    \caption{\textbf{Quantitative results of the weight of attention loss}. In practice, $\alpha$=0.2 yields the best result and is adopted in all experiments.}
    \label{tab:attention_weight}
    \vspace{-1mm}
    
    \setlength\tabcolsep{3pt}
  \scalebox{0.9}{
\begin{tabular}{lcccc}
\toprule
$\alpha$ & CLIP-T$\uparrow$ & CLIP-I$\uparrow$ & DINO-I$\uparrow$ & T. Cons.$\uparrow$ \\
\midrule
\midrule
1.0   & 0.6909  & 0.6496  & 0.3773  & 0.7812  \\
\rowcolor[rgb]{ .851,  .906,  .992} {\bf 0.2}   & \textbf{0.7075} & \textbf{0.6863} & \textbf{0.3983} & \textbf{0.7960} \\
0.01  & 0.6924  & 0.6710  & 0.3715  & 0.7886  \\
\bottomrule
\end{tabular}%
}

\end{table}

\begin{table}[t]
    \centering
   
    \caption{\textbf{Quantitative results of the negative value}. Interestingly, $\eta$=-1e-8 achieves the best result in terms of all the considered metrics. Practically, we adopt this setting in all the experiments.}
    \label{tab:negative_value}
     \vspace{-1mm}
    
    \setlength\tabcolsep{3pt}
  \scalebox{0.9}{
\begin{tabular}{lcccc}
\toprule
$\eta$ & CLIP-T$\uparrow$ & CLIP-I$\uparrow$ & DINO-I$\uparrow$ & T. Cons.$\uparrow$ \\
\midrule
\midrule
-1e-5 & 0.6786  & 0.6392  & 0.3719  & 0.7604  \\
\rowcolor[rgb]{ .851,  .906,  .992} {\bf -1e-8} & \textbf{0.7075} & \textbf{0.6863} & \textbf{0.3983} & \textbf{0.7960} \\
-1e-11 & 0.6926  & 0.6737  & 0.3613  & 0.7812  \\
\bottomrule
\end{tabular}%
}

\end{table}

\section{Conclusions}

This paper provides a novel framework CustomVideo for multi-subject driven T2V generation, powered by a simple yet effective co-occurrence and attention control mechanism.
It successfully disentangles different similar subjects and preserves the corresponding identities.
We design a flexible and efficient training process, only requiring subject images rather than subject videos. 
And during inference, a desired video can be easily generated just by providing a text prompt from the user.
We collect a comprehensive dataset CustomStudio for evaluation.
Extensive quantitative and qualitative experiments, together with a user study, demonstrates the effectiveness of our CustomVideo, outperforming SOTA methods significantly.
We hope our method could serve as a strong baseline and motivate further research on subject-driven applications, especially for multi-subject scenarios.

{
\bibliographystyle{IEEEtran}
\bibliography{ref}

\begin{thebibliography}{10}
\providecommand{\url}[1]{#1}
\csname url@samestyle\endcsname
\providecommand{\newblock}{\relax}
\providecommand{\bibinfo}[2]{#2}
\providecommand{\BIBentrySTDinterwordspacing}{\spaceskip=0pt\relax}
\providecommand{\BIBentryALTinterwordstretchfactor}{4}
\providecommand{\BIBentryALTinterwordspacing}{\spaceskip=\fontdimen2\font plus
\BIBentryALTinterwordstretchfactor\fontdimen3\font minus
  \fontdimen4\font\relax}
\providecommand{\BIBforeignlanguage}[2]{{%
\expandafter\ifx\csname l@#1\endcsname\relax
\typeout{** WARNING: IEEEtran.bst: No hyphenation pattern has been}%
\typeout{** loaded for the language `#1'. Using the pattern for}%
\typeout{** the default language instead.}%
\else
\language=\csname l@#1\endcsname
\fi
#2}}
\providecommand{\BIBdecl}{\relax}
\BIBdecl

\bibitem{wang2023modelscope}
J.~Wang, H.~Yuan, D.~Chen, Y.~Zhang, X.~Wang, and S.~Zhang, ``Modelscope
  text-to-video technical report,'' \emph{arXiv preprint arXiv:2308.06571},
  2023.

\bibitem{zeng2023make}
Y.~Zeng, G.~Wei, J.~Zheng, J.~Zou, Y.~Wei, Y.~Zhang, and H.~Li, ``Make pixels
  dance: High-dynamic video generation,'' \emph{IEEE/CVF Computer Vision and
  Pattern Recognition Conference}, pp. 8850--8860, 2024.

\bibitem{blattmann2023stable}
A.~Blattmann, T.~Dockhorn, S.~Kulal, D.~Mendelevitch, M.~Kilian, D.~Lorenz,
  Y.~Levi, Z.~English, V.~Voleti, A.~Letts \emph{et~al.}, ``Stable video
  diffusion: Scaling latent video diffusion models to large datasets,''
  \emph{arXiv preprint arXiv:2311.15127}, 2023.

\bibitem{kondratyuk2023videopoet}
D.~Kondratyuk, L.~Yu, X.~Gu, J.~Lezama, J.~Huang, R.~Hornung, H.~Adam,
  H.~Akbari, Y.~Alon, V.~Birodkar \emph{et~al.}, ``Videopoet: A large language
  model for zero-shot video generation,'' \emph{International Conference on
  Machine Learning}, 2024.

\bibitem{chen2024videocrafter2}
H.~Chen, Y.~Zhang, X.~Cun, M.~Xia, X.~Wang, C.~Weng, and Y.~Shan,
  ``Videocrafter2: Overcoming data limitations for high-quality video diffusion
  models,'' in \emph{IEEE/CVF Conference on Computer Vision and Pattern
  Recognition}, 2024, pp. 7310--7320.

\bibitem{ho2020denoising}
J.~Ho, A.~Jain, and P.~Abbeel, ``Denoising diffusion probabilistic models,''
  \emph{Advances in neural information processing systems}, vol.~33, pp.
  6840--6851, 2020.

\bibitem{song2020denoising}
J.~Song, C.~Meng, and S.~Ermon, ``Denoising diffusion implicit models,''
  \emph{International Conference on Learning Representations}, 2020.

\bibitem{lu2022dpm}
C.~Lu, Y.~Zhou, F.~Bao, J.~Chen, C.~Li, and J.~Zhu, ``Dpm-solver: A fast ode
  solver for diffusion probabilistic model sampling in around 10 steps,''
  \emph{Advances in neural information processing systems}, vol.~35, pp.
  5775--5787, 2022.

\bibitem{zhao2023videoassembler}
H.~Zhao, T.~Lu, J.~Gu, X.~Zhang, Z.~Wu, H.~Xu, and Y.-G. Jiang,
  ``Videoassembler: Identity-consistent video generation with reference
  entities using diffusion model,'' \emph{arXiv preprint arXiv:2311.17338},
  2023.

\bibitem{jiang2024videobooth}
Y.~Jiang, T.~Wu, S.~Yang, C.~Si, D.~Lin, Y.~Qiao, C.~C. Loy, and Z.~Liu,
  ``Videobooth: Diffusion-based video generation with image prompts,''
  \emph{IEEE/CVF Computer Vision and Pattern Recognition Conference}, 2024.

\bibitem{wei2024dreamvideo}
Y.~Wei, S.~Zhang, Z.~Qing, H.~Yuan, Z.~Liu, Y.~Liu, Y.~Zhang, J.~Zhou, and
  H.~Shan, ``Dreamvideo: Composing your dream videos with customized subject
  and motion,'' in \emph{IEEE/CVF Computer Vision and Pattern Recognition
  Conference}, 2024.

\bibitem{chen2023videodreamer}
H.~Chen, X.~Wang, G.~Zeng, Y.~Zhang, Y.~Zhou, F.~Han, and W.~Zhu,
  ``Videodreamer: Customized multi-subject text-to-video generation with
  disen-mix finetuning,'' \emph{arXiv preprint arXiv:2311.00990}, 2023.

\bibitem{rombach2022high}
R.~Rombach, A.~Blattmann, D.~Lorenz, P.~Esser, and B.~Ommer, ``High-resolution
  image synthesis with latent diffusion models,'' in \emph{IEEE/CVF Computer
  Vision and Pattern Recognition Conference}, 2022, pp. 10\,684--10\,695.

\bibitem{zhang2024attention}
Y.~Zhang, M.~Yang, Q.~Zhou, and Z.~Wang, ``Attention calibration for
  disentangled text-to-image personalization,'' \emph{IEEE/CVF Computer Vision
  and Pattern Recognition Conference}, 2024.

\bibitem{kirillov2023segany}
A.~Kirillov, E.~Mintun, N.~Ravi, H.~Mao, C.~Rolland, L.~Gustafson, T.~Xiao,
  S.~Whitehead, A.~C. Berg, W.-Y. Lo, P.~Doll{\'a}r, and R.~Girshick, ``Segment
  anything,'' \emph{International Conference on Computer Vision}, 2023.

\bibitem{chen2023videocrafter1}
H.~Chen, M.~Xia, Y.~He, Y.~Zhang, X.~Cun, S.~Yang, J.~Xing, Y.~Liu, Q.~Chen,
  X.~Wang \emph{et~al.}, ``Videocrafter1: Open diffusion models for
  high-quality video generation,'' \emph{arXiv preprint arXiv:2310.19512},
  2023.

\bibitem{duan2023diffsynth}
Z.~Duan, L.~You, C.~Wang, C.~Chen, Z.~Wu, W.~Qian, J.~Huang, F.~Chao, and
  R.~Ji, ``Diffsynth: Latent in-iteration deflickering for realistic video
  synthesis,'' \emph{arXiv preprint arXiv:2308.03463}, 2023.

\bibitem{balaji2019conditional}
Y.~Balaji, M.~R. Min, B.~Bai, R.~Chellappa, and H.~P. Graf, ``Conditional gan
  with discriminative filter generation for text-to-video synthesis,'' in
  \emph{International Joint Conference on Artificial Intelligence}, vol.~1, no.
  2019, 2019, p.~2.

\bibitem{skorokhodov2022stylegan}
I.~Skorokhodov, S.~Tulyakov, and M.~Elhoseiny, ``Stylegan-v: A continuous video
  generator with the price, image quality and perks of stylegan2,'' in
  \emph{IEEE/CVF Computer Vision and Pattern Recognition Conference}, 2022, pp.
  3626--3636.

\bibitem{hong2022cogvideo}
W.~Hong, M.~Ding, W.~Zheng, X.~Liu, and J.~Tang, ``Cogvideo: Large-scale
  pretraining for text-to-video generation via transformers,''
  \emph{International Conference on Learning Representations}, 2023.

\bibitem{villegas2023phenaki}
R.~Villegas, M.~Babaeizadeh, P.-J. Kindermans, H.~Moraldo, H.~Zhang, M.~T.
  Saffar, S.~Castro, J.~Kunze, and D.~Erhan, ``Phenaki: Variable length video
  generation from open domain textual descriptions,'' in \emph{International
  Conference on Learning Representations}, 2023.

\bibitem{zhu2023motionvideogan}
J.~Zhu, H.~Ma, J.~Chen, and J.~Yuan, ``Motionvideogan: A novel video generator
  based on the motion space learned from image pairs,'' \emph{IEEE Transactions
  on Multimedia}, vol.~25, pp. 9370--9382, 2023.

\bibitem{goodfellow2014generative}
I.~J. Goodfellow, J.~Pouget-Abadie, M.~Mirza, B.~Xu, D.~Warde-Farley, S.~Ozair,
  A.~Courville, and Y.~Bengio, ``Generative adversarial nets,'' in
  \emph{Advances in neural information processing systems}, 2014, pp.
  2672--2680.

\bibitem{van2017neural}
A.~van~den Oord, O.~Vinyals, and K.~Kavukcuoglu, ``Neural discrete
  representation learning,'' in \emph{Advances in neural information processing
  systems}, 2017, pp. 6309--6318.

\bibitem{he2022latent}
Y.~He, T.~Yang, Y.~Zhang, Y.~Shan, and Q.~Chen, ``Latent video diffusion models
  for high-fidelity video generation with arbitrary lengths,'' \emph{arXiv
  preprint arXiv:2211.13221}, 2022.

\bibitem{wang2023lavie}
Y.~Wang, X.~Chen, X.~Ma, S.~Zhou, Z.~Huang, Y.~Wang, C.~Yang, Y.~He, J.~Yu,
  P.~Yang \emph{et~al.}, ``Lavie: High-quality video generation with cascaded
  latent diffusion models,'' \emph{arXiv preprint arXiv:2309.15103}, 2023.

\bibitem{zhang2023show}
D.~J. Zhang, J.~Z. Wu, J.-W. Liu, R.~Zhao, L.~Ran, Y.~Gu, D.~Gao, and M.~Z.
  Shou, ``Show-1: Marrying pixel and latent diffusion models for text-to-video
  generation,'' \emph{arXiv preprint arXiv:2309.15818}, 2023.

\bibitem{zhang2023i2vgen}
S.~Zhang, J.~Wang, Y.~Zhang, K.~Zhao, H.~Yuan, Z.~Qin, X.~Wang, D.~Zhao, and
  J.~Zhou, ``I2vgen-xl: High-quality image-to-video synthesis via cascaded
  diffusion models,'' \emph{arXiv preprint arXiv:2311.04145}, 2023.

\bibitem{hu2024benchmark}
Y.~Hu, C.~Luo, and Z.~Chen, ``A benchmark for controllable text -image-to-video
  generation,'' \emph{IEEE Transactions on Multimedia}, vol.~26, pp.
  1706--1719, 2024.

\bibitem{zhao2024ta2v}
M.~Zhao, W.~Wang, T.~Chen, R.~Zhang, and R.~Li, ``Ta2v: Text-audio guided video
  generation,'' \emph{IEEE Transactions on Multimedia}, vol.~26, pp.
  7250--7264, 2024.

\bibitem{koksal2023controllable}
A.~K{\"o}ksal, K.~E. Ak, Y.~Sun, D.~Rajan, and J.~H. Lim, ``Controllable video
  generation with text-based instructions,'' \emph{IEEE transactions on
  multimedia}, vol.~26, pp. 190--201, 2023.

\bibitem{singer2023makeavideo}
U.~Singer, A.~Polyak, T.~Hayes, X.~Yin, J.~An, S.~Zhang, Q.~Hu, H.~Yang,
  O.~Ashual, O.~Gafni, D.~Parikh, S.~Gupta, and Y.~Taigman, ``Make-a-video:
  Text-to-video generation without text-video data,'' in \emph{International
  Conference on Learning Representations}, 2023.

\bibitem{blattmann2023align}
A.~Blattmann, R.~Rombach, H.~Ling, T.~Dockhorn, S.~W. Kim, S.~Fidler, and
  K.~Kreis, ``Align your latents: High-resolution video synthesis with latent
  diffusion models,'' in \emph{IEEE/CVF Computer Vision and Pattern Recognition
  Conference}, 2023, pp. 22\,563--22\,575.

\bibitem{khachatryan2023text2video}
L.~Khachatryan, A.~Movsisyan, V.~Tadevosyan, R.~Henschel, Z.~Wang,
  S.~Navasardyan, and H.~Shi, ``Text2video-zero: Text-to-image diffusion models
  are zero-shot video generators,'' \emph{International Conference on Computer
  Vision}, 2023.

\bibitem{guo2023animatediff}
Y.~Guo, C.~Yang, A.~Rao, Y.~Wang, Y.~Qiao, D.~Lin, and B.~Dai, ``Animatediff:
  Animate your personalized text-to-image diffusion models without specific
  tuning,'' \emph{International Conference on Learning Representations}, 2024.

\bibitem{wang2023videocomposer}
X.~Wang, H.~Yuan, S.~Zhang, D.~Chen, J.~Wang, Y.~Zhang, Y.~Shen, D.~Zhao, and
  J.~Zhou, ``Videocomposer: Compositional video synthesis with motion
  controllability,'' \emph{Advances in Neural Information Processing Systems},
  vol.~36, 2024.

\bibitem{lin2023videodirectorgpt}
H.~Lin, A.~Zala, J.~Cho, and M.~Bansal, ``Videodirectorgpt: Consistent
  multi-scene video generation via llm-guided planning,'' \emph{arXiv}, 2023.

\bibitem{masala2023explaining}
M.~Masala, N.~Cudlenco, T.~Rebedea, and M.~Leordeanu, ``Explaining vision and
  language through graphs of events in space and time,'' in \emph{International
  Conference on Computer Vision}, 2023, pp. 2826--2831.

\bibitem{wu2023tune}
J.~Z. Wu, Y.~Ge, X.~Wang, S.~W. Lei, Y.~Gu, Y.~Shi, W.~Hsu, Y.~Shan, X.~Qie,
  and M.~Z. Shou, ``Tune-a-video: One-shot tuning of image diffusion models for
  text-to-video generation,'' in \emph{International Conference on Computer
  Vision}, 2023.

\bibitem{qi2023fatezero}
C.~Qi, X.~Cun, Y.~Zhang, C.~Lei, X.~Wang, Y.~Shan, and Q.~Chen, ``Fatezero:
  Fusing attentions for zero-shot text-based video editing,''
  \emph{International Conference on Computer Vision}, 2023.

\bibitem{geyer2023tokenflow}
M.~Geyer, O.~Bar-Tal, S.~Bagon, and T.~Dekel, ``Tokenflow: Consistent diffusion
  features for consistent video editing,'' \emph{International Conference on
  Learning Representations}, 2024.

\bibitem{gal2023an}
R.~Gal, Y.~Alaluf, Y.~Atzmon, O.~Patashnik, A.~H. Bermano, G.~Chechik, and
  D.~Cohen-or, ``An image is worth one word: Personalizing text-to-image
  generation using textual inversion,'' in \emph{International Conference on
  Learning Representations}, 2023.

\bibitem{kumari2023multi}
N.~Kumari, B.~Zhang, R.~Zhang, E.~Shechtman, and J.-Y. Zhu, ``Multi-concept
  customization of text-to-image diffusion,'' in \emph{IEEE/CVF Computer Vision
  and Pattern Recognition Conference}, 2023, pp. 1931--1941.

\bibitem{avrahami2023break}
O.~Avrahami, K.~Aberman, O.~Fried, D.~Cohen-Or, and D.~Lischinski,
  ``Break-a-scene: Extracting multiple concepts from a single image,'' in
  \emph{SIGGRAPH Asia}, 2023, pp. 1--12.

\bibitem{ma2023subject}
J.~Ma, J.~Liang, C.~Chen, and H.~Lu, ``Subject-diffusion: Open domain
  personalized text-to-image generation without test-time fine-tuning,'' in
  \emph{ACM SIGGRAPH 2024 Conference Papers}, 2024, pp. 1--12.

\bibitem{jiang2024animediff}
Y.~Jiang, Q.~Liu, D.~Chen, L.~Yuan, and Y.~Fu, ``Animediff: Customized image
  generation of anime characters using diffusion model,'' \emph{IEEE
  Transactions on Multimedia}, 2024.

\bibitem{liu2023cones}
Z.~Liu, R.~Feng, K.~Zhu, Y.~Zhang, K.~Zheng, Y.~Liu, D.~Zhao, J.~Zhou, and
  Y.~Cao, ``Cones: Concept neurons in diffusion models for customized
  generation,'' \emph{International Conference on Machine Learning}, 2023.

\bibitem{liu2023cones2}
Z.~Liu, Y.~Zhang, Y.~Shen, K.~Zheng, K.~Zhu, R.~Feng, Y.~Liu, D.~Zhao, J.~Zhou,
  and Y.~Cao, ``Cones 2: Customizable image synthesis with multiple subjects,''
  \emph{Advances in neural information processing systems}, 2023.

\bibitem{ruiz2023dreambooth}
N.~Ruiz, Y.~Li, V.~Jampani, Y.~Pritch, M.~Rubinstein, and K.~Aberman,
  ``Dreambooth: Fine tuning text-to-image diffusion models for subject-driven
  generation,'' in \emph{IEEE/CVF Computer Vision and Pattern Recognition
  Conference}, 2023, pp. 22\,500--22\,510.

\bibitem{xiao2023fastcomposer}
G.~Xiao, T.~Yin, W.~T. Freeman, F.~Durand, and S.~Han, ``Fastcomposer:
  Tuning-free multi-subject image generation with localized attention,''
  \emph{arXiv preprint arXiv:2305.10431}, 2023.

\bibitem{wei2024mm}
Z.~Wei, Q.~Su, L.~Qin, and W.~Wang, ``Mm-diff: High-fidelity image
  personalization via multi-modal condition integration,'' \emph{arXiv preprint
  arXiv:2403.15059}, 2024.

\bibitem{jiang2024mc}
J.~Jiang, Y.~Zhang, K.~Feng, X.~Wu, and W.~Zuo, ``Mc$^2$: Multi-concept
  guidance for customized multi-concept generation,'' \emph{arXiv preprint
  arXiv:2404.05268}, 2024.

\bibitem{he2024id}
X.~He, Q.~Liu, S.~Qian, X.~Wang, T.~Hu, K.~Cao, K.~Yan, M.~Zhou, and J.~Zhang,
  ``Id-animator: Zero-shot identity-preserving human video generation,''
  \emph{arXiv preprint arXiv:2404.15275}, 2024.

\bibitem{hu2022lora}
E.~J. Hu, yelong shen, P.~Wallis, Z.~Allen-Zhu, Y.~Li, S.~Wang, L.~Wang, and
  W.~Chen, ``Lo{RA}: Low-rank adaptation of large language models,'' in
  \emph{International Conference on Learning Representations}, 2022.

\bibitem{zeroscope}
S.~Sterling, ``Zeroscope,''
  \url{https://huggingface.co/cerspense/zeroscope_v2_576w}, 2023.

\bibitem{claude3}
Anthropic, ``Introducing the next generation of claude,''
  \url{https://www.anthropic.com/news/claude-3-family}, 2024.

\bibitem{gu2023mix}
Y.~Gu, X.~Wang, J.~Z. Wu, Y.~Shi, Y.~Chen, Z.~Fan, W.~Xiao, R.~Zhao, S.~Chang,
  W.~Wu \emph{et~al.}, ``Mix-of-show: Decentralized low-rank adaptation for
  multi-concept customization of diffusion models,'' \emph{Advances in Neural
  Information Processing Systems}, vol.~36, 2023.

\bibitem{loshchilov2017decoupled}
I.~Loshchilov and F.~Hutter, ``Decoupled weight decay regularization,''
  \emph{arXiv preprint arXiv:1711.05101}, 2017.

\bibitem{ho2022classifier}
J.~Ho and T.~Salimans, ``Classifier-free diffusion guidance,'' in \emph{NeurIPS
  2021 Workshop on Deep Generative Models and Downstream Applications}, 2021.

\bibitem{zeroscopexl}
S.~Sterling, ``Zeroscope xl,''
  \url{https://huggingface.co/cerspense/zeroscope_v2_XL}, 2023.

\bibitem{paszke2019pytorch}
A.~Paszke, S.~Gross, F.~Massa, A.~Lerer, J.~Bradbury, G.~Chanan, T.~Killeen,
  Z.~Lin, N.~Gimelshein, L.~Antiga \emph{et~al.}, ``Pytorch: An imperative
  style, high-performance deep learning library,'' \emph{Advances in neural
  information processing systems}, vol.~32, 2019.

\bibitem{diffusers2022}
P.~von Platen, S.~Patil, A.~Lozhkov, P.~Cuenca, N.~Lambert, K.~Rasul,
  M.~Davaadorj, and T.~Wolf, ``Diffusers: State-of-the-art diffusion models,''
  \url{https://github.com/huggingface/diffusers}, 2022.

\bibitem{accelerate2022sylvain}
S.~Gugger, L.~Debut, T.~Wolf, P.~Schmid, Z.~Mueller, S.~Mangrulkar, M.~Sun, and
  B.~Bossan, ``Accelerate: Training and inference at scale made simple,
  efficient and adaptable,'' \url{https://github.com/huggingface/accelerate},
  2022.

\bibitem{radford2021learning}
A.~Radford, J.~W. Kim, C.~Hallacy, A.~Ramesh, G.~Goh, S.~Agarwal, G.~Sastry,
  A.~Askell, P.~Mishkin, J.~Clark \emph{et~al.}, ``Learning transferable visual
  models from natural language supervision,'' in \emph{International Conference
  on Machine Learning}, 2021, pp. 8748--8763.

\bibitem{dosovitskiy2021an}
A.~Dosovitskiy, L.~Beyer, A.~Kolesnikov, D.~Weissenborn, X.~Zhai,
  T.~Unterthiner, M.~Dehghani, M.~Minderer, G.~Heigold, S.~Gelly, J.~Uszkoreit,
  and N.~Houlsby, ``An image is worth 16x16 words: Transformers for image
  recognition at scale,'' in \emph{International Conference on Learning
  Representations}, 2021.

\bibitem{caron2021emerging}
M.~Caron, H.~Touvron, I.~Misra, H.~J{\'e}gou, J.~Mairal, P.~Bojanowski, and
  A.~Joulin, ``Emerging properties in self-supervised vision transformers,'' in
  \emph{International Conference on Computer Vision}, 2021, pp. 9650--9660.

\bibitem{esser2023structure}
P.~Esser, J.~Chiu, P.~Atighehchian, J.~Granskog, and A.~Germanidis, ``Structure
  and content-guided video synthesis with diffusion models,'' in
  \emph{International Conference on Computer Vision}, 2023, pp. 7346--7356.

\end{thebibliography}
}

\end{document}